\documentclass[lettersize,journal]{IEEEtran}
\usepackage{amsmath,amsfonts}
\usepackage{algorithmic}
\usepackage{array}
\usepackage[caption=false,font=normalsize,labelfont=sf,textfont=sf]{subfig}
\usepackage{textcomp}
\usepackage{stfloats}
\usepackage{url}
\usepackage{verbatim}
\usepackage{graphicx}
\usepackage{cite}

\usepackage{multirow}
\usepackage{makecell}
\usepackage[ruled,linesnumbered]{algorithm2e}
\usepackage{soul}
\usepackage{color, xcolor}
\usepackage{colortbl}
\usepackage{booktabs}
\usepackage{amssymb}

\begin{document}

\title{Unleashing the Potential of Tracklets for \\Unsupervised Video Person Re-Identification}

\author{Nanxing Meng$^{\dagger}$, Qizao Wang$^{\dagger}$, Bin Li, Xiangyang Xue

\thanks{Nanxing Meng, Qizao Wang, Bin Li, and Xiangyang Xue are with the School of Computer Science, and Shanghai Key Lab of Intelligent Information Processing, Fudan University, China. Email: \{nxmeng21, qzwang22\}@m.fudan.edu.cn, \{libin, xyxue\}@fudan.edu.cn.
Bin Li is the corresponding author.
}
\thanks{${\dagger}$ These authors contributed equally to this work.}
}

\markboth{IEEE TRANSACTIONS ON INFORMATION FORENSICS AND SECURITY}{}


\maketitle

\begin{abstract}
With rich temporal-spatial information, video-based person re-identification methods have shown broad prospects. Although tracklets can be easily obtained with ready-made tracking models, annotating identities is still expensive and impractical. Therefore, some video-based methods propose using only a few identity annotations or camera labels to facilitate feature learning. They also simply average the frame features of each tracklet, overlooking unexpected variations and inherent identity consistency within tracklets. In this paper, we propose the Self-Supervised Refined Clustering (SSR-C) framework without relying on any annotation or auxiliary information to promote unsupervised video person re-identification. 
Specifically, we first propose the Noise-Filtered Tracklet Partition (NFTP) module to reduce the feature bias of tracklets caused by noisy tracking results, and sequentially partition the noise-filtered tracklets into ``sub-tracklets''. Then, we cluster and further merge sub-tracklets using the self-supervised signal from the tracklet partition, which is enhanced through a progressive strategy to generate reliable pseudo labels, facilitating intra-class cross-tracklet aggregation. Moreover, we propose the Class Smoothing Classification (CSC) loss to efficiently promote model learning. Extensive experiments on the MARS and DukeMTMC-VideoReID datasets demonstrate that our proposed SSR-C for unsupervised video person re-identification achieves state-of-the-art results and is comparable to advanced supervised methods. The code is available at \url{https://github.com/Darylmeng/SSRC-Reid}.
\end{abstract}

\begin{IEEEkeywords}
Person re-identification, unsupervised learning, video tracklets, reliable clustering.
\end{IEEEkeywords}

\section{Introduction}
\label{sec:intro}

\IEEEPARstart{P}{erson} Re-IDentification (Re-ID) aims to retrieve a query identity from a gallery of disjoint cameras \cite{zheng2016person,ye2021deep}. In recent years, person Re-ID has made great progress with the prosperity of deep neural networks~\cite{qian2019leader,he2021transreid,wang2022co,wang2024exploring,wang2025content,wang2025distribution,wang2025image}. Compared with image-based methods~\cite{zhao2016person,sun2018beyond,zhou2019omni,shu2021large,wang2023rethinking}, video-based methods~\cite{yan2020learning,yang2020spatial,wu2020adaptive,hou2021bicnet,zang2022exploiting,hou2019vrstc,wu2019progressive,zeng2022anchor} usually achieve better performance with the help of rich temporal-spatial information. 
Nonetheless, prevailing video-based approaches predominantly rely on supervised learning, necessitating costly cross-camera tracklet annotations, thereby impeding their broader deployment in practical real-world settings.
Given the substantial challenges and impracticalities associated with obtaining a wealth of annotated tracklets, alternative strategies have emerged: \textbf{(1)} One-shot based methods~\cite{wu2018exploit,ye2017dynamic,liu2017stepwise,ye2018robust} propose to label only one tracklet for each identity; \textbf{(2)} Tracklet association based methods~\cite{li2019unsupervised,chen2018deep,zang2022exploiting,zeng2022anchor} explore the correspondence of within-camera and cross-camera tracklets, which assumes the camera labels are available. 
However, these methods still require several pedestrian tracklet annotations or camera information, thereby restricting their universal applicability in diverse real-world scenarios.

\begin{figure}[t]
\centering
  \includegraphics[width=1\linewidth]{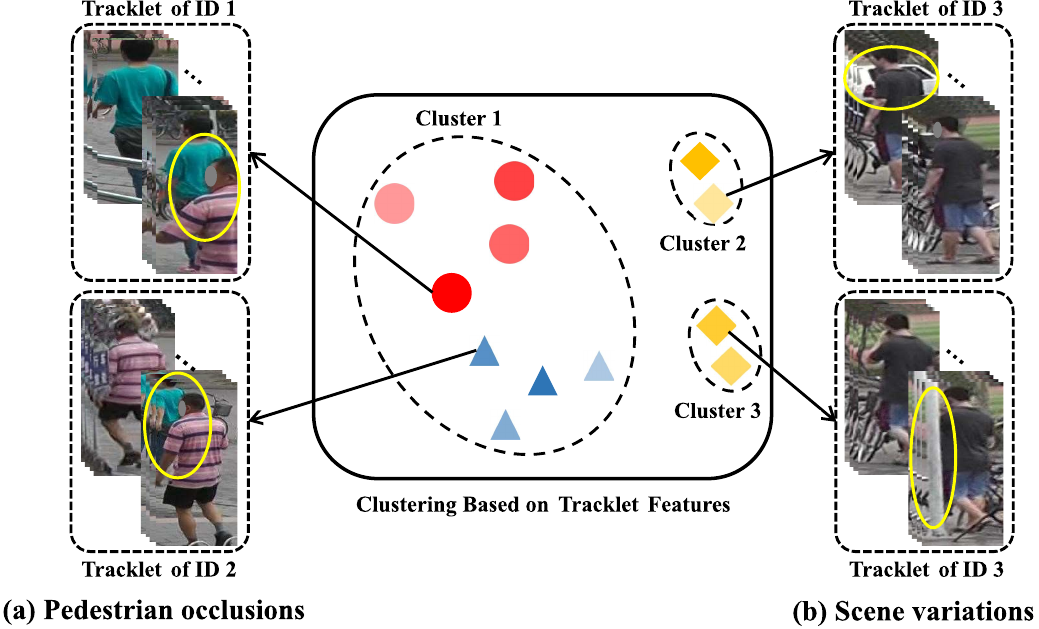}
  \caption{\textbf{Limitations of directly applying clustering on tracklets to obtain pseudo labels.} Different shapes denote different identities, and the same shapes with various colors represent different tracklets of the same pedestrian. Due to the noises and unexpected variations within tracklets (marked in yellow ovals), (a) different individuals are wrongly clustered, and (b) tracklets of the same person are incorrectly pushed away. }
  \label{fig:intro}
\end{figure}

Drawing inspiration from established practices in unsupervised image-based Re-ID methods~\cite{li2022cluster,ge2020self,cho2022part,chen2021ice,lan2023learning}, a more practical and feasible solution is to perform the clustering algorithm to generate pseudo identity labels for training. 
The straightforward implementation is to conduct clustering directly based on the tracklet features. However, since tracklets are obtained using off-the-shelf tracking models in an unsupervised manner, noisy frames caused by tracking errors are inevitable, and some difficult frames, \textit{e.g.}, with heavy occlusions or appearance variations, can also introduce bias to the tracklet features. 

On the one hand, as illustrated in Fig.~\ref{fig:intro}~(a), with the tracklet overlapping between pedestrians, the tracklet features of different individuals may be close in the embedding space, so that different identities are incorrectly assigned the same pseudo label. On the other hand, as shown in Fig.~\ref{fig:intro}~(b), due to the presence of occlusions and background variations, tracklets of the same identity may be mistakenly assigned to different clusters.
The noises and unexpected variations of tracklet features would weaken the representativeness and discriminability of tracklets, undermining their capacity to accurately depict distinct identities and reducing their effectiveness in distinguishing between identities. As a result, imprecise clustering results can lead to noisy pseudo labels to misguide the learning of the model.
Additionally, we notice that \textit{the noises within a tracklet and the similarity between tracklets exhibit locality}. For example, the overlapping of the two pedestrians only exists in some of their whole tracklets in Fig.~\ref{fig:intro} (a), and there are similar visual features in local tracklets in Fig.~\ref{fig:intro} (b). It is worth exploring and taking advantage of this locality to help refine clustering results, promoting effective feature learning.

Therefore, we first propose the Noise-Filtered Tracklet Partition (NFTP) module. It filters noisy frames within each tracklet in an adaptive manner and then sequentially partition each tracklet into multiple segments, referred to as ``sub-tracklets''. 
Based on the noise-filtered sub-tracklets, we propose the self-supervised refined clustering to generate pseudo labels effectively. To improve the quality of generated pseudo labels, we first acquire reliable sub-clusters via performing clustering under strict restrictions, and then merge sub-clusters into ultimate clusters.
More specifically, we make use of the identity consistency between sub-tracklets, that is, sub-tracklets from the same original tracklet share the same identity, to merge sub-clusters. In this process, due to the inherent local temporal-spatial similarity between tracklets, noise-filtered sub-tracklets would act as the intra-class relays to aggregate different cross-camera tracklets of the same pedestrian.

As our method does not rely on annotations or auxiliary information, it inevitably encounters clustering inaccuracies, which can exert a notable impact on the model's learning process. Of particular concern is that during the early stages of training, the model's poor feature extraction capabilities may exacerbate the prevalence of such clustering mistakes. Additionally, as training goes on, the model becomes more robust and the extracted features become more reliable.
Therefore, we propose a progressive merging strategy to alleviate the impact of inevitable clustering inaccuracies on model learning, which improves the discriminative ability and robustness of the model effectively. 
Eventually, with reliable pseudo labels and sub-tracklets with less feature bias, we propose the Class Smoothing Classification (CSC) loss to effectively guide the model learning at a more fine-grained sub-tracklet level.
By iteratively performing our proposed self-supervised refined clustering and effective feature learning, the capability and robustness of the model can be significantly enhanced. \textbf{Our contributions can be summarized as follows:}

\textbf{(1)} We propose a novel unsupervised video person Re-ID framework based on self-supervised refined clustering, without using any additional information. It unleashes the potential of tracklets with the help of noise-filtered sub-tracklets, effectively reducing clustering errors and enhancing model capability.

\textbf{(2)} We propose to take advantage of the identity consistency within tracklets and treat sub-tracklets as intra-class relays to facilitate intra-class cross-tracklet aggregation, resulting in reliable clustering results. Furthermore, we introduce a progressive merging strategy and the class smoothing classification loss at the sub-tracklet level to promote feature learning.

\textbf{(3)} Our proposed method achieves state-of-the-art results on the MARS and DukeMTMC-VideoReID datasets for video-based person Re-ID, surpassing other unsupervised methods by a large margin, and achieves comparable results with advanced supervised methods.

\section{Related Work}
\subsection{Unsupervised Video Person Re-Identification}
To alleviate the difficulty of labeling identities, early unsupervised video person Re-ID methods are based on one-shot learning~\cite{ye2017dynamic,liu2017stepwise,wu2018exploit,ye2018robust,wu2019progressive}, which requires at least one annotated tracklet for each person identity, and technically they are semi-supervised. For instance, Liu et al.~\cite{liu2017stepwise} use each training tracklet as a query, and perform retrieval in the cross-camera training set. 
Together with the initial labeled data, Wu et al.~\cite{wu2018exploit} propose a progressive sampling method to increase the number of the selected pseudo-labeled candidates step by step to generate an enlarged training set.
However, relying on one labeled tracklet for each identity still restricts their flexibility in real-world applications. To eliminate the need for identity annotations, existing methods for unsupervised video person Re-ID can be broadly categorized into two types: tracklet association based methods and clustering-based methods.

\subsubsection{Tracklet association based methods} Of this type, research tends to investigate tracklet association across different cameras \cite{chen2018deep,li2019unsupervised,zang2022exploiting,zeng2022anchor}. For example,
Li et al.~\cite{li2019unsupervised} simultaneously conduct tracklet association learning within-camera and across-camera, as well as a soft labeling approach to mitigate penalties for potentially positive samples. 
Chen~et~al.~\cite{chen2018deep} explore local space-time consistency within each tracklet from the same camera view and global cyclic ranking consistency between tracklets across disjoint camera views. Zang et al.~\cite{zang2022exploiting} also explore the potential of local-level features for unsupervised learning.
However, the reliance on additional camera information renders them inflexible in real-world applications where these clues may be inaccessible.

\subsubsection{Clustering-based methods \label{subsec:clustering_related_work}} These methods provide a more effective way for unsupervised person Re-ID~\cite{fan2018unsupervised,lin2019bottom,wu2020tracklet,xie2021unsupervised,ding2019towards,lin2020unsupervised,dai2022cluster,xie2022sampling,wang2023relation}. They adopt clustering for pseudo identity labels, which are used for model optimization. Fan et al.~\cite{fan2018unsupervised} propose a progressive unsupervised learning method that iterates between pedestrian clustering with K-means~\cite{macqueen1967some} and fine-tuning the model. However, it requires a pre-defined number of clusters, which can be impractical in real-world applications.
Bottom-Up Clustering (BUC)~\cite{lin2019bottom} progressively incorporates samples into clusters, but negative samples can be inevitably included, impacting the quality of the generated pseudo labels. Following BUC, some works alleviate the adverse impact of noisy frames during clustering~\cite{xie2021unsupervised} or introduce a dispersion-based criterion to achieve high-quality clustering~\cite{ding2019towards}. 
Wu et al.~\cite{wu2020tracklet} design a comprehensive unsupervised learning objective that accounts for tracklet frame coherence, tracklet neighborhood compactness, and tracklet cluster structure.
Dai et al.~\cite{dai2022cluster} employ a memory dictionary and computes contrastive loss at the cluster level.
Xie et al.~\cite{xie2022sampling} resample and re-weight the hard frames in tracklets to improve clustering ability.
Tao~et~al.~\cite{tao2024unsupervised} perform a denoising diffusion process~\cite{ho2020denoising} and guide global feature learning in patch context to refine pseudo labels.
However, due to the inherent suboptimal clustering robustness of these methods, it is difficult for them to achieve satisfactory performance.

\subsection{Self-Supervised Learning for Person Re-Identification}
Self-supervised learning aims at utilizing the intrinsic characteristics of data by designing unsupervised auxiliary supervisory signal for feature learning~\cite{tang2019unsupervised,jiang2020self,wu2020tracklet,dou2023identity}. In the context of person Re-ID, it plays a crucial role due to the inherent difficulty of manual annotation in this task.
For example, Tang et al.~\cite{tang2019unsupervised} use the model itself to create supervision on the iterative pseudo-labeling process. 
A popular self-supervised learning method in person Re-ID is contrastive learning~\cite{hermans2017defense}. It involves creating positive and negative pairs, and training the model to maximize similarity between positives while minimizing it between negatives.
With this paradigm, Jiang et al.~\cite{jiang2020self} design a triplet loss to model the self-supervised constraint of data samples in cross domain, and Wu et al.~\cite{wu2020tracklet} perform self-supervised tracklet frame coherence constraint based on augmented frames.
Dou et al.~\cite{dou2023identity} construct positive pairs from inter-frame images by modeling the instance association as a maximum-weight bipartite matching problem and further present a reliability-guided contrastive loss to suppress the adverse impact of noisy positive pairs.
Different from existing methods, we take advantage of the intrinsic identity consistency signal to perform self-supervised refined clustering, so as to obtain more accurate pseudo labels and facilitate effective feature learning.

\section{Methodology}
We first formulate the unsupervised video person Re-ID task in Sec.~\ref{subsec:formulation}. Then, we provide the overview of the clustering-based baseline that is based on the original tracklets in Sec.~\ref{subsec:overview}, where we extend the image-based work~\cite{hu2021hard} to the video scenario as our baseline. Considering harmful factors within tracklets, we propose noise-filtered tracklet partition strategy, which results in sub-tracklets (See Sec.~\ref{subsec:NFTP}). Based on sub-tracklets, we further introduce self-supervised refined clustering in Sec.~\ref{subsec:SSR}, resulting in abundant pseudo labels of sub-tracklets. After that, to optimize the model effectively, we introduce an effective loss in Sec.~\ref{subsec:csc}.
Finally, we introduce the training details of our proposed clustering-based framework in Sec.~\ref{subsec:training}.

\subsection{Task Formulation}
\label{subsec:formulation}
Given an unlabeled training set $X = \left \{ T_{1},T_{2},\cdots,T_{N}\right \} $ of $N$ tracklets, where each tracklet contains $L_{i}$ frames, \textit{i.e.}, $T_{i} =  \left \{ x_{i,j}  \right \} _{j=1}^{L_{i}} $, the goal is to optimize the model $\phi$ to extract discriminative identity feature from each input tracklet $T_{i}$. 
During inference, the optimized model $\phi$ is used to extract features of tracklets from the query and gallery sets, respectively. Then, the features of the query tracklets are used to search the gallery set and retrieve the matched tracklets according to the cosine distance metric.

\begin{figure*}[t]
\centering
  \includegraphics[width=0.93\linewidth]{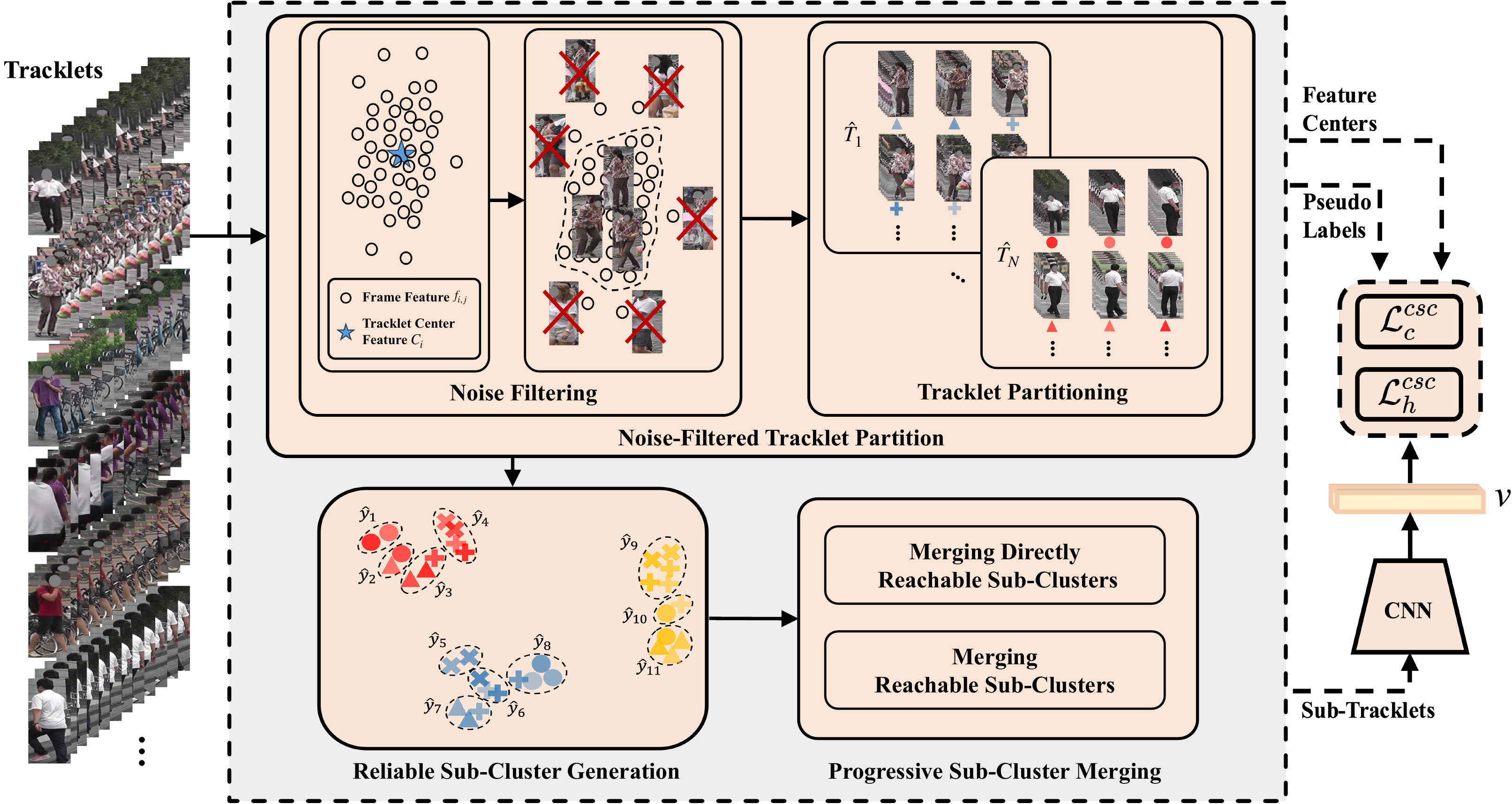}
  \caption{\textbf{The framework of our proposed SSR-C.} 
  During each training epoch, our proposed Noise-Filtered Tracklet Partition (NFTP) module would filter noisy frames within each tracklet, and further sequentially partition each tracklet into multiple sub-tracklets. Different shapes (\textit{e.g.}, triangle and circle) correspond to different tracklets, different colors (\textit{e.g.}, red and blue) denote different identities, and various transparencies of the same color represent the sub-tracklets of each tracklet. With noise-filtered sub-tracklets, we can acquire reliable sub-clusters via strictly restricted clustering, and then sub-clusters are merged leveraging the self-supervised signal from tracklet partitioning and following a progressive merging strategy. Finally, we design the Class Smoothing Classification (CSC) loss, which leverages the obtained accurate pseudo labels to effectively improve the discriminative ability of the model. }
  \label{fig:pipeline}
\end{figure*}

\subsection{Clustering-based Framework Overview}
\label{subsec:overview}
\subsubsection{Pseudo Label Acquisition} During the training process, $M$ frames of each tracklet are randomly selected and average pooled to figure out the feature embedding $\boldsymbol{v}_{i} $ for each tracklet. Formally,
\begin{equation}
\boldsymbol{v}_{i} = \frac{1}{M}\sum_{j=1}^{M} \phi \left (\theta ;x_{i,j} \right ),
\label{eq:tracklet_feat}
\end{equation}
where $\boldsymbol{v}_{i} \in \mathbb{R}^{d}$, and $\phi$ is the feature embedding model parameterized by $\theta $. For notation simplicity, we denote $\phi \left (\theta ;x_{i,j} \right )$ as $f_{i,j} \in \mathbb{R}^{d}$. 
Since the identification annotations of the tracklets are inaccessible in the unsupervised setting, a density-based clustering method DBSCAN~\cite{ester1996density} is performed to obtain pseudo labels, which does not require prior knowledge of the total number of identities. 

After applying DBSCAN, the clustered tracklets are assigned to the same pseudo label and isolated outliers are simply discarded. Then, we can get a ``labeled'' dataset $\hat{X} = \{ \left ( T_{i},\hat{y_{i}}\right )  \}_{i=1}^{N}$, where $1\le \hat{y_{i}} \le n$ denotes the pseudo label of the $i$-th tracklet, and $n$ is the total number of clusters.
Additionally, we calculate the centroid of each cluster by averaging tracklet features within it to initialize a memory $\boldsymbol{V} \in \mathbb{R}^{n \times d}$.

\subsubsection{Training} At each training step, based on the pseudo labels and the memory, InfoNCE~\cite{oord2018representation} loss is adopted for model optimization. Formally,
\begin{align}
\mathcal{L}_{c} &= -\log \left(p\left(\hat{y}_{i} \mid T_{i}, \boldsymbol{V} \right)\right) \\
&= -\log \frac{\exp \left( \boldsymbol{v_{i}}^{\top} \cdot \boldsymbol{V}_{\hat{y_{i}}} / \tau \right)}{\sum_{j=1}^{n} \exp \left( \boldsymbol{v_{i}}^{\top} \cdot \boldsymbol{V}_{j} / \tau \right)},
\label{repelledloss}
\end{align}
where $\boldsymbol{v}_{i}$ and $\hat{y_{i}}$ are the feature embedding and pseudo label of the $i$-th tracklet, $\boldsymbol{V}_{j}$ is the $j$-th entry of $\boldsymbol{V}$, and $\tau$ is the temperature parameter~\cite{hinton2015distilling} to control the softness of the probability distribution over classes. 
During back-propagation, we update $\boldsymbol{V}$ with the momentum update strategy as follows:
\begin{equation}
\boldsymbol{V}_{j}\gets \alpha \boldsymbol{V}_{j}+\left ( 1-\alpha  \right ) \overline{\boldsymbol{v}}_{j},
\label{eq:memoryupdate}
\end{equation}
where $\overline{v}_{j}$ is the average of tracklet features with the pseudo label $j$ in the mini-batch.

To further distinguish easily confused sample pairs and improve the inter-class separability, similar to~\cite{hu2021hard}, we also maintain an additional hard memory $\boldsymbol{V}^{h}$ storing the hard tracklet feature of each class. 
$\boldsymbol{V}^{h}$ is initialized the same as $\boldsymbol{V}$ and is updated in a similar way to Eq. \ref{eq:memoryupdate}, except that we mine the hardest sample in the mini-batch for the hard memory update. Specifically, we compute the cosine similarity between each sample with pseudo label $\hat{y}_{i}$ in the mini-batch and the corresponding memory entry $\boldsymbol{V}_{\hat{y_{i}}}^{h}$, and the hardest sample for class $\hat{y}_{i}$ is the one with the lowest cosine similarity.
The overall loss function is formulated as:
\begin{equation}
\mathcal{L}_{h} = -\log \left(p\left(\hat{y}_{i} \mid T_{i}, \boldsymbol{V}^{h} \right)\right),
\label{L_h}
\end{equation}
\begin{equation}
\mathcal{L} = \gamma _{1}\mathcal{L}_{h} + \gamma _{2}\mathcal{L}_{c},
\label{eq:lossfunc}
\end{equation}
where Eq.~\ref{L_h} is defined similar to Eq.~\ref{repelledloss}, $\gamma _{1}$ and $\gamma _{2}$ are hyper-parameters to balance between $\mathcal{L}_{h}$ and $\mathcal{L}_{c}$. By iteratively performing clustering and model training, the feature extraction capability of the model as well as the accuracy of the obtained pseudo labels could be gradually enhanced.

\subsection{Noise-Filtered Tracklet Partition}
\label{subsec:NFTP}
Performing clustering on tracklet features can generate inaccurate pseudo labels since there are numerous harmful frames and unexpected variations within tracklets. Thus, we propose the Noise-Filtered Tracklet Partition (NFTP) module to eliminate these negative impacts, further promoting our self-supervised refined clustering.

\subsubsection{Noise Filtering}
\label{subsec:NF}
At the beginning of each epoch, we use the latest model to extract the frame features $\left \{ f_{i,j} \right \} _{j=1}^{L_{i}} $ of each tracklet $T_{i}$, and then average them to obtain the center feature $C_{i}$ as follows:
\begin{equation}
    C_{i} = \frac{1}{L_{i}}\sum_{j=1}^{L_{i}}f_{i,j}.
\label{eq:center_feat}
\end{equation}

Intuitively, noisy frames will deviate more from the center of the tracklet than other frames. Therefore, inspired by \cite{xie2021unsupervised}, we apply an adaptive threshold to filter out noisy frames, which is computed as:
\begin{equation}
\begin{aligned}
q_{i} = \frac{1}{L_{i}*\delta } \sum_{j=1}^{L_{i}} {\rm dist} \left( f_{i,j}, C_{i} \right),   \\
\end{aligned}
\label{eq:filteringthreshold}
\end{equation}
where $ {\rm dist} \left( f_{i,j}, C_{i} \right)$ is computed as ${\left( 1 - f_{i,j} \cdot C_{i} \right)}^{2}$ and $f_{i,j} \cdot C_{i}$ denotes the cosine similarity between each frame $f_{i,j}$ and the center feature $C_{i}$. $\delta$ is a hyperparameter that adjusts the degree of noise filtering.
For each frame $x_{i,j}$, if $ {\rm dist} \left( f_{i,j}, C_{i} \right) > q_{i}$, it is considered a noisy frame. The threshold $q_{i}$ is dynamically adjusted according to the frame features and the length of tracklets, so that noises within tracklets can be effectively filtered. 
To help understand the effectiveness of the design of dynamic noise filtering, we provide more discussions on the impact of $\delta$ on adaptive threshold and the visualization of noisy frames in Figs.~\ref{fig:2param} and \ref{fig:noise-filter}, respectively.

\subsubsection{Tracklet Partition}
\label{subsec:TP}
To alleviate unexpected variations within tracklets, given each noise-filtered tracklet $\hat{T}_{i} =  \left \{ x_{i,j}  \right \} _{j=1}^{\hat{L}_{i}} $, where $\hat{L}_{i}$ is the number of frames, we employ a sequentially equidistant partition strategy to divide each tracklet $\hat{T}_{i}$ into multiple ``sub-tracklets''. 
Formally,
\begin{equation}
    \hat{T}_{i} = \left \{ \hat{ST}_{i,t} \right \}_{t=1}^{n'_{i}}, 
    ~ \hat{ST}_{i,t}=\{ x_{i,j} \}^{t \cdot l}_{j=1+(t-1)l},
\label{eq:partition}
\end{equation}
where $n'_{i}$ represents the number of the sub-tracklets from tracklet $\hat{T}_{i}$, and $l$ is the fixed partition stride.
After partition, the unlabeled training dataset can be re-formulated as $X = \left \{ \hat{ST}_{1}, \hat{ST}_{2},\cdots, \hat{ST}_{N^{s}}\right \} $, where $N^{s} = \sum_{i=1}^{N}{n'_{i}}$ is the total number of sub-tracklets.

\subsection{Self-Supervised Refined Clustering}
\label{subsec:SSR}
Based on the noised-filtered sub-tracklets, we propose the self-supervised refined clustering to obtain accurate pseudo labels. As shown in Fig. \ref{fig:pipeline}, it first generates reliable sub-clusters and then merges them in a self-supervised manner to aggregate cross-camera tracklets of the same identity.

\subsubsection{Reliable Sub-Cluster Generation}
DBSCAN \cite{ester1996density} is employed for clustering, which utilizes a scanning radius hyper-parameter, denoted as $eps$, to regulate the size of the cluster. To ensure the consistency of identities within clusters, we set a strict constraint condition of small $eps$. In this way, only more reliable sub-tracklets, \textit{i.e.}, sub-tracklets with relatively close feature distance with each other, are considered to share the same identity. After clustering, each cluster is assigned with a different pseudo label $\hat{y}$, and we can get the labeled dataset $\hat{X} = \{  ( \hat{ST}_{i}, \hat{y_{i}} ) \}_{i=1}^{N^{s}}$, where $\hat{y_{i}}$ denotes the pseudo label of the $i$-th sub-tracklet.
Since our proposed NFTP module has filtered noises within sub-tracklets, the distance between sub-tracklets of different identities is extended. They are harder to be clustered under the strict clustering constraint, alleviating the problem in Fig. \ref{fig:intro} (a).
However, there are still spatial-temporal changes in each tracklet, such as complicated backgrounds or illumination variations, so sub-tracklets of the same tracklet can also be grouped into different clusters under such a strict clustering process. Therefore, we further merge the obtained clusters, which are viewed as ``sub-clusters'', to form the ultimate clustering results.

\subsubsection{Progressive Sub-Cluster Merging}
The natural self-supervised signal from tracklet partitioning in Sec.~\ref{subsec:TP} can serve as a guide for sub-cluster merging. Since sub-tracklets are obtained through the sequential partition, various sub-tracklets from the same tracklet indicate the same identity. Additionally, as observed in Fig. \ref{fig:intro}, cross-camera tracklets may contain similar local areas, so that cross-camera sub-tracklets of the same identity can be clustered into the same sub-cluster. 
With the merging process based on the self-supervised signal from the tracklet partition, various cross-camera tracklets of the same identity can be effectively aggregated together. 

\begin{figure}[t]
\centering
  \includegraphics[width=0.93\linewidth]{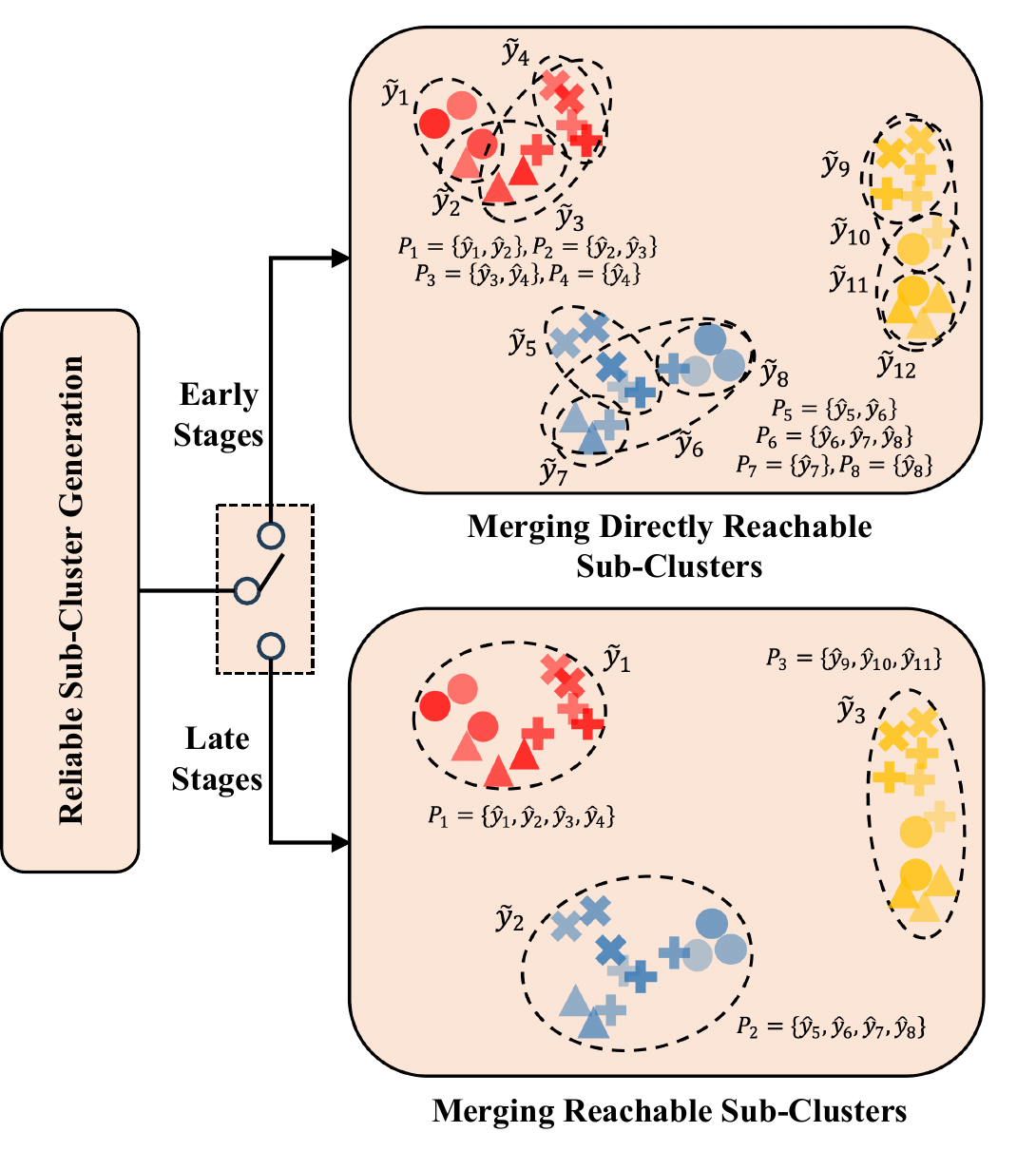}
  \caption{\textbf{Illustration of our proposed progressive sub-cluster merging.} Different shapes (\textit{e.g.}, triangle and circle) correspond to different tracklets, different colors (\textit{e.g.}, red and blue) denote different identities, and various transparencies of the same color represent the sub-tracklets of each tracklet. Based on the generated reliable sub-clusters, only directly reachable sub-clusters are merged during the early stages of training, while all reachable sub-clusters are merged later.
  }
  \label{fig:PM}
\end{figure}

Considering the feature extraction capability of the model is inferior at the early training stages, and improves as training progresses, we incorporate a progressive strategy to facilitate effective sub-cluster merging as shown in Fig. \ref{fig:PM}.  
For a clear description, we define two terms here:

\newtheorem{definition}{Definition}
\begin{definition}
\textit{Given a tracklet $T_{A}$, if its sub-tracklets $ST_{A}$ are assigned to several sub-clusters $\left \{ c_{a},c_{b},\cdots,c_{Z} \right \}$, then these $Z$ sub-clusters are considered \textbf{Directly Reachable}~(D.R.) with each other.}
\label{def1}
\end{definition}

\begin{definition}
\textit{Given tracklets $T_{A}$ and $T_{B}$, if sub-tracklets $ST_{A}$ are assigned to sub-clusters $c_{a}$ and $c_{b}$, and sub-tracklets $ST_{B}$ are assigned to sub-clusters $c_{b}$ and $c_{c}$, then the two sub-clusters $c_{a}$ and $c_{c}$ are considered \textbf{Reachable}~(R.).}
\label{def2}
\end{definition}

During the early stages of training, due to the poor feature extraction ability, sub-tracklets from different identities may be assigned to the same sub-cluster. To avoid error propagation during the merging step, we only merge directly reachable sub-clusters (Def.~\ref{def1}) at the early training stages. As training goes on, the model becomes more robust and sub-clusters become more reliable. Therefore, we propose to merge all reachable sub-clusters (Def.~\ref{def2}) after the learning rate decay. 
Benefiting from the progressive sub-cluster merging strategy, reliable clusters can effectively promote feature learning of the model, making it less affected by clustering noises.

After sub-cluster merging, each merged cluster is assigned with a different refined pseudo label (denoted by $\tilde{y}$), and all the sub-tracklets in each merged cluster share the same refined pseudo label. Finally, with our proposed self-supervised refined clustering, we can get the dataset $\tilde{X} = \{ ( \hat{ST}_{i}, \hat{y_{i}}, \tilde{y_{i}} ) \}_{i=1}^{N^{s}}$ with two kinds of pseudo labels for each sub-tracklet. As introduced in Sec.~\ref{subsec:csc}, the abundant pseudo labels $\hat{y}$ and $\tilde{y}$ enable effective optimization in the fine-grained sub-tracklet embedding space.

\begin{algorithm}[t]
\small
\caption{The training procedure of SSR-C}
\label{algorithm}
\KwIn{Unlabeled training set  $X = \left \{ T_{1},T_{2},\cdots,T_{N}\right \} $; Number of tracklets $N$; CNN model $\phi \left ( \theta  \right )$; Total training epoch $EP$; Training iterations per epoch $ITER$; Training batch size $B$.
}
\For{epoch $\gets$ {\rm 1 to} $EP$}{
    Freeze the model $\phi \left ( \theta  \right )$\;
    \tcc{Perform the NFTP module.}
    \For{$i$ $\gets$ {\rm 1 to} $N$}{
    $C_{i}$ $\gets$ Compute the center feature of $T_{i}$ via Eq.~\ref{eq:center_feat}\;
    $\hat{T}_{i}$ $\gets$ Filter noisy frames in $T_{i}$ via Eq. \ref{eq:filteringthreshold}\;
    $\hat{ST}$ $\gets$ Partition $\hat{T}_{i}$ into sub-tracklets via Eq.~\ref{eq:partition}\;
    }
    $X \gets \{ \hat{ST}_{1}, \hat{ST}_{2},\cdots, \hat{ST}_{N^{s}} \} $\;
    
    \tcc{Perform self-supervised refined clustering.}
    \For{$i$ $\gets$ {\rm 1 to} $N^{s}$}{
    $v_{i} \gets$ Compute the feature of $\hat{ST}_{i}$ via Eq. \ref{eq:tracklet_feat}\;
    }
    $\hat{y} \gets$ Perform reliable sub-cluster generation\;
    $\tilde{y} \gets$ Perform progressive sub-cluster merging\;
    $\tilde{X} \gets \{ ( \hat{ST}_{i}, \hat{y_{i}}, \tilde{y_{i}} ) \}_{i=1}^{N^{s}}$\;
    Initialize $\boldsymbol{V}$ and $\boldsymbol{V}^{h}$ with sub-cluster feature centers\;
    
    \tcc{Train the model with $\tilde{X}$.}
    Unfreeze the model $\phi \left ( \theta \right )$\;
    \For{iter $\gets$ {\rm 1 to} $ITER$}{
        $\{\hat{ST}_{i}\}_{i=1}^{B}$ $\gets$ Sample a training batch\;
        \For{$i$ $\gets$ {\rm 1 to} $B$}{
        $\boldsymbol{v}_{i} \gets$ Extract the feature of $\hat{ST}_{i}$ via Eq. \ref{eq:tracklet_feat}\;
        Optimize $\theta$ via Eqs.~\ref{eq:csc_loss}, \ref{eq:labeldistribution} and \ref{eq:lossfunc}\;
        }
        Update $\boldsymbol{V}$ and $\boldsymbol{V}^{h}$ with
sub-tracklet features via Eq.~\ref{eq:memoryupdate}\;
    }
}
\KwOut{Optimal $\theta$.}
\end{algorithm}

\subsection{Class Smoothing Classification Loss}
\label{subsec:csc}
With abundant pseudo labels of sub-tracklets, we design an effective loss, namely Class Smoothing Classification (CSC) loss, to effectively guide the model learning at the fine-grained sub-tracklet level.
As shown in Figs.~\ref{fig:pipeline} and \ref{fig:PM}, after adopting our self-supervised refined clustering, each merged cluster (assigned with refined pseudo label $\tilde{y}$) is composed of multiple sub-clusters (assigned with pseudo label $\hat{y}$). Therefore, each refined pseudo label $\tilde{y}_{i}$ correspond to more than one $\hat{y}$. We denote the set of all $\hat{y}$ corresponding to each $\tilde{y_{i}}$ as $P_{i}$ (as shown in Fig.~\ref{fig:PM}), and define the CSC loss as follows:
{
\begin{equation}
\resizebox{0.98\hsize}{!}{$
\mathcal{L}_{c}^{csc} \! = - \! \! \!~ \sum_{j \in P_{i}} s_{j} \log \frac{\exp \left( \boldsymbol{v_{i}}^{\top} \! \cdot \! \boldsymbol{V}_{j} / \tau\right)}{\exp \left( \boldsymbol{v_{i}}^{\top} \! \cdot \! \boldsymbol{V}_{j} / \tau\right) \! + \! \! \! \!
\sum\limits_{k \notin P_{i}} \exp \left( \boldsymbol{v_{i}}^{\top} \! \cdot \! \boldsymbol{V}_{k} / \tau\right)},
$}
\label{eq:csc_loss}
\end{equation}
}
\begin{equation}
s_{j} = \left\{\begin{array}{lr}
1 - \lambda + \frac{\lambda}{K},  &j=\hat{y}_{i} \\
\frac{\lambda}{K}, &{\rm otherwise}
\end{array}\right. 
, \quad \forall j \in P_{i},
\label{eq:labeldistribution} 
\end{equation}
where $\lambda$ is the smoothing term, $K$ is the cardinal number of the positive set $P_{i}$, and $\boldsymbol{V}$ is the memory, which is initialized with sub-cluster feature centers and updated using sub-tracklet features following Eq. \ref{eq:memoryupdate}. Similarly, we can define $\mathcal{L}_{h}^{csc}$ based on $\boldsymbol{V^{h}}$.

We note that some works~\cite{li2019unsupervised,lin2020unsupervised,zhong2019invariance,zhong2020learning} simply select $K$ potential positive samples to mitigate the impact of clustering errors. They adopt feature similarity as metrics and treat $K$ as a predefined hyper-parameter, which limits their methods in two ways. 
\textbf{(1)} The feature extraction ability of the model is changing at different training stages, making it difficult to determine the number of positive samples. There are also various positive samples for different people, so a \textbf{static} $K$ makes lots of samples mistakenly optimized. 
\textbf{(2)} Some hard potential positive samples would not be identified simply based on feature similarity. In contrast, our method effectively addresses these problems by determining the value of $K$ \textbf{dynamically} based on both feature similarities and guidance of self-supervised signals.
With the empowerment of our proposed NFTP module and self-supervised refined clustering, accurate $K$ positive sub-clusters can guide the learning of the model in the fine-grained sub-tracklet embedding space, improving its discriminative ability.

Our proposed CSC loss also exhibits significant differences from the widely-used label smoothing regularization~\cite{szegedy2016rethinking}. Formally, we set small non-zero values for the weight term (\textit{i.e.}, $s_{j}$ in Eq.~\ref{eq:labeldistribution}) of potential positive classes, but the weight terms of negative classes are still 0. In contrast, label smoothing regularization sets the weight terms of all negative classes to small non-zero values. Moreover, CSC loss aims to optimize the model in a more fine-grained sub-tracklet embedding space leveraging our refined pseudo labels $\hat{y_{i}}$ and $\tilde{y_{i}}$, while label smoothing regularization is used to avoid overfitting.

\subsection{Training Details of SSR-C}
\label{subsec:training}
Generally, we follow the training framework as introduced in Sec. \ref{subsec:overview}, except that our proposed SSR-C optimizes the model at the sub-tracklet level. The training procedure of our proposed SSR-C framework is provided in Alg.~\ref{algorithm}.
At the beginning of each epoch, we first freeze the model and apply our proposed NFTP module to obtain noisy-filtered sub-tracklets. Then, we extract the features of noisy-filtered sub-tracklets via Eq. \ref{eq:tracklet_feat}, with the only difference that we randomly sample $M$ frames for each noisy-filtered sub-tracklet rather than original tracklet. Subsequently, our proposed self-supervised refined clustering method is conducted based on the sub-tracklet features. After that, the memories $\boldsymbol{V}$ and $\boldsymbol{V^{h}}$ are initialized with sub-cluster feature centers, and updated with sub-tracklet features via Eq.~\ref{eq:memoryupdate}. Meanwhile, our proposed CSC loss is used to guide the model learning following Eq. \ref{eq:lossfunc}.
By iteratively performing our proposed self-supervised refined clustering with NFTP and effective model optimization, the feature extraction capability and robustness of the model can be enhanced.

\section{Experiments}

\subsection{Datasets}
\label{subsec:dataset}
We evaluate our method on two widely used video person Re-ID datasets, \textit{i.e.}, MARS~\cite{zheng2016mars} and DukeMTMC-VideoReID \cite{wu2018exploit}.
\textbf{MARS} is captured by six cameras on a university campus. It contains 17,503 tracklets for 1,261 distinct identities and 3,248 distractor tracklets, which are divided into 625 identities for training and 636 identities for testing. 
\textbf{DukeMTMC-VideoReID} (Duke-V) is derived from DukeMTMC~\cite{ristani2016performance}, which is captured in outdoor scenes with noisy backgrounds and suffers from illumination, pose, and viewpoint changes and occlusions. 
Duke-V is split following the protocol~\cite{zheng2017unlabeled}, with 2,196 tracklets of 702 identities for training and 2,636 tracklets of other 702 identities for testing.

\subsection{Evaluation Metrics}
The Cumulative Matching Characteristic (CMC)~\cite{wang2007shape} at Rank-1/5/10, and the mean Average Precision (mAP)~\cite{market1501} are used for performance evaluation.

\begin{table*}[t]
\centering
\caption{\textbf{Comparison with state-of-the-art methods on two video person Re-ID datasets, \textit{i.e.}, MARS and Duke-V.} ``Label'' denotes requiring one tracklet annotation for each person. ``Cameras'' denotes requiring camera labels. Supervised methods using manually annotated identity labels are represented by ``${\dagger}$''. The best results are shown in bold.}
\label{comparsion}
\setlength{\tabcolsep}{3.7mm}{
\begin{tabular}{lcccccccccc}
\toprule
{\multirow{2}{*}{\textbf{Methods}}} & \multirow{2}{*}{\textbf{Year}} & \multirow{2}{*}{\begin{tabular}[c]{@{}c@{}}\textbf{Extra}\\ \textbf{Info.}\end{tabular}} & \multicolumn{4}{c}{\textbf{MARS}} & \multicolumn{4}{c}{\textbf{Duke-V}}\\
\cmidrule(r){4-7} \cmidrule(r){8-11}
& & & mAP & Rank-1 & Rank-5 & Rank-10 & mAP & Rank-1 & Rank-5 & Rank-10 \\
\midrule

\rowcolor{gray!10} {VRSTC$^{\dagger}$~\cite{hou2019vrstc}} & 2019	&	&82.3	&88.5	&96.5	&97.4	&93.5	&95.0	&99.1	&99.4\\
\rowcolor{gray!10} {AGRL$^{\dagger}$~\cite{wu2020adaptive}} & 2020 &	&81.9	&89.5	&96.6	&-	&95.4	&97.0	&99.3	&-\\
\rowcolor{gray!10} {STGCN$^{\dagger}$~\cite{yang2020spatial}} & 2020	& Labels &83.7	&89.9	&96.4	&-	&95.7	&97.3	&99.3	&-\\
\rowcolor{gray!10} {AGNet$^{\dagger}$~\cite{zhao2022gait}} & 2022	&	&84.3	&89.4	&96.7	&-	&96.0	&96.3	&99.4	&-\\
\rowcolor{gray!10} {GPNet$^{\dagger}$~\cite{pan2023multi}} & 2023	&	&85.1 &90.2 &96.8	&-	&96.1 &96.3 &99.6	&-\\
\midrule

{DGM+IDE~\cite{ye2017dynamic}} 	& 2017 &  \multirow{6}{*}{Label}	&16.9	&36.8	&54.0	&-	&33.6	&42.4	&57.9	&68.3\\ 
{Stepwise~\cite{liu2017stepwise}} 	& 2017 &	&19.6	&41.2	&55.5	&-	&46.7	&56.2	&70.3	&79.2\\ 
{Race~\cite{ye2018robust}}	& 2018 & 	&24.5	&43.2	&57.1	&62.1	&-	&-	&-	&-\\ 
{EUG~\cite{wu2018exploit}}	& 2018 &	&42.4	&62.6	&74.9	&-	&63.2	&72.7	&84.1	&-\\ 
{ProLearn~\cite{wu2019progressive}} 	& 2019 &	&42.6	&62.8	&75.2	&80.4	&63.3	&72.9	&84.3	&88.3\\ 
{SPUE-Net~\cite{zhang2023self}} 	& 2023 &	&46.3	&66.7	&78.1	&81.6	&65.0	&73.3	&84.8	&89.0\\ 
\midrule

{DAL~\cite{chen2018deep}}	& 2018& \multirow{4}{*}{Cameras} &23.0	&49.3	&65.9	&72.2	&-	&-	&-	&-\\ 
{UTAL~\cite{li2019unsupervised}}	& 2019 &	&35.2	&49.9	&66.4	&-	&-	&-	&-	&-\\ 
{uPMnet~\cite{zang2022exploiting}}	& 2022 &	&-	&-	&-	&-	&76.9	&83.6	&93.1	&-\\
{UAAL~\cite{zeng2022anchor}}	& 2022 &	&60.1	&73.2	&86.3	&89.5	&87.0	&89.7	&97.0	&98.6\\ 
\midrule

{OIM~\cite{xiao2017joint}}	& 2017 & \multirow{10}{*}{None} &13.5	&33.7	&48.1	&54.8	&43.8	&51.1	&70.5	&76.2\\
{TSSL~\cite{wu2020tracklet}}	& 2020 &	&30.5	&56.3	&-	&-	&64.6	&73.9	&-	&-\\
{DBC~\cite{ding2019towards}}	& 2019 &	&31.7	&58.5	&70.1	&73.5	&67.4	&75.6	&88.5	&91.0\\
{BUC~\cite{lin2019bottom}}	& 2019 &  &38.0	&61.1	&75.1	&80.0	&61.9	&69.2	&81.1	&85.8\\
{NHAC~\cite{xie2021unsupervised}} 	& 2021 &	&40.1	&61.8	&75.3	&79.9	&76.0	&82.8	&92.7	&95.6\\
{SSL~\cite{lin2020unsupervised}} 	& 2020 &	&43.6	&62.8	&77.2	&80.1	&69.3	&76.4	&88.7	&91.0\\
{RPE~\cite{wang2023relation}} 	& 2023 &	&40.4	&63.3	&75.4	&80.6	&71.5	&77.8	&89.3	&91.7\\
{SRC~\cite{xie2022sampling}} 	& 2022 &	&40.5	&62.7	&76.1	&80.0	&76.5	&83.0	&93.3	&95.0\\
NPSSL~\cite{wang2025noise} & 2025 & & - & - & - & - & 88.9 & 88.7 & 98.1 & - \\
DBMPCL~\cite{zhang2025dual} & 2025 & & 71.6 & 82.0 & 91.1 & 93.4 & 92.7 & 94.2 & 98.7 & 99.4  \\
\midrule

SSR-C (Ours)	& 2025 & None	&\textbf{78.7}	&\textbf{85.6}	&\textbf{93.0}	&\textbf{95.1}	&\textbf{95.3}	&\textbf{95.9}	&\textbf{99.2}	&\textbf{99.6}\\
\bottomrule

\end{tabular}}
\end{table*}

\subsection{Implementation Details}
We adopt ResNet-50~\cite{he2016deep} pre-trained on ImageNet~\cite{krizhevsky2017imagenet} as our backbone model and use GeM pooling~\cite{radenovic2018fine} to obtain the feature embedding of each frame.
During training, we set the batch size to 32 and total training epochs to 150. Each tracklet is sequentially partitioned every $l=32$ frames to obtain sub-tracklets. For each sub-tracklet, we randomly sample $M=8$ frames with a stride of 4 to form a sub-tracklet segment. 
Images are resized to $256\times128$ and augmented by horizontal flipping and random erasing~\cite{zhong2020random}. As for the optimizer, AdamW with a weight decay of 0.0005 is adopted to update the parameters. The learning rate is initialized to $3.5\times 10^{-4} $ with a decay factor of 0.1 every 50 epochs. 
At the beginning of each epoch, we use DBSCAN~\cite{ester1996density} and Jaccard distance~\cite{zhong2017re} for clustering with the $eps$ set to 0.25 and the minimal number of samples for each cluster set to 2. For hyper-parameters, $\lambda$ and $\delta$ are set to 0.1 and 0.7, respectively. Following~\cite{hu2021hard}, we simply set $\tau$, $\alpha$, $\gamma_{1}$, and $\gamma_{2}$ to 0.05, 0.1, 0.5, and 0.25, respectively.

\subsection{Comparison with the State-of-the-Art}
In Tab.~\ref{comparsion}, we present a comparative analysis of our proposed SSR-C against state-of-the-art methods on two large-scale video-based person Re-ID datasets. Relatively speaking, one-shot learning based and tracklet association based methods usually achieve better performance as shown in Tab.~\ref{comparsion}, attributed to their utilization of a few identity annotations or camera metadata for guidance.
However, our proposed SSR-C still shows great superiority over them in performance, without relying on any auxiliary information. For example, on MARS, our proposed method significantly outperforms the state-of-the-art one-shot learning based method SPUE-Net~\cite{zhang2023self}, which applies a cooperative learning of local uncertainty estimation combined with determinacy estimation, and improves mAP by 32.4\%.
SSR-C also shows great superiority to the advanced tracklet association based method UAAL~\cite{zeng2022anchor}, which uses the association ranking method to discover positive and negative tracklets pairs for triplet loss training, with an improvement of 18.6\% in mAP.

Our proposed SSR-C surpasses all state-of-the-art unsupervised methods, which overlook not only the unexpected noises and variations within tracklets but also abundant useful identity information in tracklets, by a large margin without whistles and bells. 
For instance, SSR-C surpasses SRC~\cite{xie2022sampling}, which resamples and re-weights the hard frames in tracklets to improve clustering ability, with 18.8\%/12.9\% mAP/Rank-1 on Duke-V, and surpasses SSL~\cite{lin2020unsupervised}, which utilizes re-assigned soft label distribution to learn from similar images with smooth constraint, with 35.1\%/22.8\% mAP/Rank-1 on MARS.

It is worth noting that, without expensive annotations or cumbersome modules, our method is competitive with advanced supervised methods~\cite{wu2020adaptive,hou2019vrstc,yang2020spatial,zhao2022gait,pan2023multi} on both datasets. 
Recently, NPSSL~\cite{wang2025noise} devises cluster-level and sample-level filters to jointly exploit pseudo labels, while DBMPCL~\cite{zhang2025dual} explores the discriminant features of diversity distribution in both frame-level tracklet modeling and tracklet-level discrepant contrastive learning. Differently, we perform refined clustering and model optimization at the sub-tracklet level, leveraging the property of local similarity between tracklets. The significant performance gains of SSR-C demonstrate its effectiveness in generating reliable pseudo labels and enabling more discriminative feature learning.

\begin{table*}[t]
\centering
\caption{\textbf{Ablation studies on the MARS and Duke-V datasets.} ``SSR'' denotes our self-supervised refined clustering, which first generates reliable sub-clusters and then merges them self-supervised. ``R./D.R.'' and ``PM'' refer to merging reachable/directly reachable sub-clusters throughout training and the progressive sub-cluster merging strategy, respectively.}
\setlength{\tabcolsep}{4.6mm}{
\begin{tabular}{lcccccccccc}
\toprule
\multirow{2}{*}{\textbf{Methods}} & \multirow{2}{*}{\textbf{NFTP}} & \multicolumn{3}{c}{\textbf{SSR}} & \multirow{2}{*}{\textbf{InfoNCE}} & \multirow{2}{*}{\textbf{CSC}} & \multicolumn{2}{c}{\bf MARS} & \multicolumn{2}{c}{\bf Duke-V} \\ 
\cmidrule(r){3-5} \cmidrule(r){8-9} \cmidrule(r){10-11}
& & {{R.}} & {{D.R.}} & {{PM}} & & & mAP        & Rank-1       & mAP         & Rank-1       \\ 
\midrule

1 (Baseline) &                      &                     &                       &                     & $\checkmark$                        &                      & 63.9       & 75.4         & 3.2         & 4.0          \\
2 & $\checkmark$ &                     &                       &                     & $\checkmark$                        &                      & 74.3       & 82.2         & 91.7        & 93.6         \\
\rowcolor{gray!10}3 & $\checkmark$ & $\checkmark$                   &                       &                     & $\checkmark$                        &                      & 72.9       & 80.4         & 94.0        & 94.4         \\
\rowcolor{gray!10}4 & $\checkmark$ &                     & $\checkmark$                     &                     & $\checkmark$                        &                      & 72.5       & 80.2         & 94.2        & 94.5         \\
5 & $\checkmark$ & $\checkmark$                   &                       &                     &                          & $\checkmark$                    & 78.3       & 84.5         & 94.8        & 95.4         \\
6 & $\checkmark$ &                     & $\checkmark$                     &                     &                          & $\checkmark$                    & 77.6       & 84.1         & 95.0        & 95.4         \\ \midrule

7 (Ours) & $\checkmark$ &                     &                       & $\checkmark$                   &                          & $\checkmark$                    & \textbf{78.7}       & \textbf{85.6}         & \textbf{95.3}        & \textbf{95.9}         \\ 
\bottomrule
\end{tabular}}
\label{moduelablation}
\end{table*}

\begin{table}[t]
\centering
\caption{\textbf{Comparison with image-based methods on the MARS and Duke-V datasets.} All methods are replicated using DBSCAN for clustering. ``Cameras'' denotes requiring camera labels. Our proposed tracklet partitioning is adopted for comparison methods on Duke-V to ensure convergence.}
\setlength{\tabcolsep}{1.4mm}{
    \begin{tabular}{lcccccc}
    \toprule
    {\multirow{2}{*}{\textbf{Methods}}} & \multirow{2}{*}{\textbf{Year}} & \multirow{2}{*}{\begin{tabular}[c]{@{}c@{}}\textbf{Extra}\\ \textbf{Info.}\end{tabular}} & \multicolumn{2}{c}{\textbf{MARS}} & \multicolumn{2}{c}{\textbf{Duke-V}} \\ 
    \cmidrule(r){4-5} \cmidrule(r){6-7}
    & & & mAP   & Rank-1  & mAP    & Rank-1 \\ \midrule
    
    ICE~\cite{chen2021ice} & 2021 & \multirow{2}{*}{Cameras} & 72.3  & 81.0 & 86.6   & 88.2 \\
    IIDCL~\cite{xiong2024inter} & 2024 & & 61.2 & 74.2 & 88.0 & 88.9 \\
    \midrule
    
    SPCL~\cite{ge2020self} & 2020 & \multirow{3}{*}{None} & 44.7  & 59.5  & 77.0   & 79.5 \\
    PPLR~\cite{cho2022part} & 2022 & & 53.6  & 65.9  & 86.1    & 86.2    \\
    Cluster Contrast~\cite{dai2022cluster} & 2022 & & 54.4 & 67.8 & 85.5 & 86.5 \\
    \midrule
    
    SSR-C (Ours) & 2025 & None & \textbf{78.7}  & \textbf{85.6} & \textbf{95.3}   & \textbf{95.9} \\ 
    \bottomrule
    \end{tabular}}
\label{image_compare}
\end{table}

\subsubsection{More Comparison with Image-based Methods}
In Tab.~\ref{image_compare}, we further compare with state-of-the-art image-based unsupervised person Re-ID methods by using DBSCAN for clustering and replicating their results on the video-based MARS and Duke-V datasets. Note that we adopt tracklet partitioning for all these image-based methods on Duke-V, otherwise, they struggle to converge as observed in Tab.~\ref{moduelablation} (see detailed discussion in Sec.~\ref{subsec:ablation}). 
SPCL~\cite{ge2020self} improves clustering via hybrid memory and reliability criterion, PPLR~\cite{cho2022part} reduces the pseudo label noise by employing the complementary relationship between global and part features, and Cluster Contrast~\cite{dai2022cluster} computes contrastive loss at the cluster level.
Differently, our proposed SSR-C leverages self-supervised refined clustering to generate reliable pseudo labels and performs CSC loss for optimization at the sub-tracklet level. SSR-C surpasses them significantly on MARS and Duke-V without whistles and bells.

Moreover, SSR-C surpasses methods that utilizes additional camera information~\cite{chen2021ice,xiong2024inter,tao2024unsupervised} by a large margin on both benchmarks. Among them, ICE~\cite{chen2021ice} leverages inter-instance pairwise similarity scores and camera information to boost contrastive learning, and IIDCL~\cite{xiong2024inter} employs camera proxies to discern the disparities and similarities among individuals across inter-camera views.
Without relying on extra information, SSR-C outperforms the state-of-the-art image-based method IIDCL~\cite{xiong2024inter} by 17.5\%/11.4\% mAP/Rank-1 on MARS and 7.3\%/7.0\% mAP/Rank-1 on Duke-V, respectively. Our proposed SSR-C leverages the locality of the noises within a tracklet and the similarity between tracklets, effectively addressing the challenges posed by noises and unexpected variations within tracklets.

It is also worth mentioning that since we do not modify the backbone model or introduce any extra burden modules, our proposed SSR-C has no extra cost during inference compared with the baseline method. SSR-C is efficient in real-world application. During training, when processing all the 8,298 tracklets on MARS, our proposed NFTP module and progressive sub-cluster merging strategy only take less than 9 seconds on one NVIDIA GeForce GTX 1080 Ti.

\subsection{Ablation Studies}
\label{subsec:ablation}
In this section, we first investigate the effectiveness of each component in our proposed SSR-C framework through detailed ablation experiments in Tabs.~\ref{moduelablation}, \ref{nf_tp_ablation}, \ref{sampleamount} and \ref{pooling_abalation}. Then, to help understand our proposed method and demonstrate its superiority, we conduct in-depth analyses of different values of hyper-parameters and their impacts in Figs.~\ref{fig:3param} and \ref{fig:2param}, and compare the clustering statistics between the baseline and our method in Fig.~\ref{fig:statistic}.

\subsubsection{Effectiveness of NFTP}
Taking the noises and unexpected variations within tracklets into account, our proposed NFTP module enables reliable clustering. As shown in Tab.~\ref{moduelablation}, Method 2 with NFTP brings 10.4\% mAP improvement over Baseline (Method 1) on MARS, and a more remarkable performance improvement is observed on Duke-V. However, directly applying InfoNCE loss based on the original tracklets fails to converge. 
We argue that tracklets in Duke-V are much longer than MARS (167.6 \textit{v.s.} 59.5 frames per tracklet on average), which implies more severe variations in tracklets. If we simply average frames within each tracklet without NFTP, the tracklet features with bias would result in inaccurate pseudo labels, and further cause the degradation of the optimization. In contrast, partitioning each tracklet yields sub-tracklets with relatively mild feature changes, which facilitates the feature-based clustering algorithm to obtain accurate pseudo labels. 
Given that applying NFTP may also increase the amount of training data, we further investigate its influence in Tab.~\ref{sampleamount}.

\subsubsection{Effectiveness of CSC Loss and Self-Supervised Refined Clustering} 
Based on the abundant pseudo labels generated with our self-supervised refined clustering, we design CSC loss for effective optimization with dynamically mined positive samples. 
As shown in the shaded area in Tab.~\ref{moduelablation}, compared with Method 2, incorporating either R. or D.R. sub-cluster merging strategy and training with InfoNCE loss show inferior results on MARS. The reason is that although our proposed self-supervised refined clustering produces reliable optimization guidance, InfoNCE still optimizes the model in the coarse tracklet granularity after violent sub-cluster merging. Besides, fusing sub-clusters can inevitably involve negative identities, especially when the capability of the model is weak at the early training stage. 
However, as the results of Methods 5 and 6 show, CSC loss can take full advantage of the abundant supervision $\hat{y}$ and $\tilde{y}$ from self-supervised refined clustering. The two complement and promote each other, enhancing the discriminative ability of the model significantly.

\subsubsection{Effectiveness of Progressive Sub-Cluster Merging Strategy} 
Compared with Methods 5 and 6, Ours (Method 7) yields an additional improvement on both datasets. The results demonstrate the effectiveness of our progressive sub-cluster merging strategy in adapting the optimization to the feature extraction capability of the model at different training stages.

\begin{table}[t]
\centering
\caption{\textbf{Influence of noise filtering (NF) and tracklet partitioning (TP) in the NFTP module.}
}
\setlength{\tabcolsep}{3.8mm}{
\begin{tabular}{lcccc}
\toprule
\multirow{2}{*}{\textbf{Methods}} & \multicolumn{2}{c}{\textbf{MARS}} & \multicolumn{2}{c}{\textbf{Duke-V}} \\ \cmidrule(r){2-3} \cmidrule(r){4-5}
                         & mAP   & Rank-1  & mAP    & Rank-1  \\ \midrule
Baseline & 63.9 & 75.4 & 3.2 & 4.0 \\
Baseline $+$ NF            & 71.4  & 80.7    & 7.3 & 9.5    \\
Baseline $+$ TP            & 72.5  & 79.8    & 87.4   & 89.0\\
Baseline $+$ NFTP          & 74.3  & 82.2    & 91.7   & 93.6\\
SSR-C $-$ NF             & 73.8  & 81.9  & 93.8   & 94.6\\ \midrule
SSR-C & \textbf{78.7} & \textbf{85.6} & \textbf{95.3} & \textbf{95.9}\\ \bottomrule
\end{tabular}}
\label{nf_tp_ablation}
\vspace{-0.1in}
\end{table}

\subsubsection{Ablation Studies of Noise Filtering and Tracklet Partitioning}
We provide more detailed ablation results of our proposed NFTP module, which contains two key designs of noise filtering (NF) and tracklet partitioning (TP).
\textbf{(1)} NF is to handle harmful frames as depicted in Fig.~\ref{fig:intro}~(a). In unsupervised learning, noisy frames in tracklets can lead to inaccurate clustering and wrong pseudo labels, thereby hurting model learning. More intuitively, adopting noisy filtering to the baseline in Tab.~\ref{nf_tp_ablation} shows its effectiveness.
\textbf{(2)} TP is designed to alleviate unexpected variations within tracklets as depicted in Fig.~\ref{fig:intro}~(b). Moreover, it helps explore locality within tracklets and enables our progressive self-supervised refined clustering for reliable pseudo labels. We only adopt tracklet partitioning to the baseline and SSR-C in Tab.~\ref{nf_tp_ablation}, and the results show its non-negligible necessity.
\textbf{(3)} To promote learning and avoid the potential loss of complementary features within tracklets, we propose a progressive sub-cluster merging strategy to merge sub-tracklets of the same identity effectively. We have shown its effectiveness in Tab.~\ref{moduelablation}.

\begin{table}[t]
\centering
\caption{\textbf{The impact of the different numbers of sampled frames per tracklet (\#Frames).}}
\setlength{\tabcolsep}{2.2mm}{
\begin{tabular}{lcccccc}
\toprule
\multirow{2}{*}{\textbf{Methods}} & \multirow{2}{*}{\textbf{\#Frames}} & \multicolumn{2}{c}{\textbf{MARS}} & \multicolumn{2}{c}{\textbf{Duke-V}} \\ 
\cmidrule(r){3-4} \cmidrule(r){5-6}
& & mAP &Rank-1 & mAP & Rank-1 \\ \midrule
\multirow{2}{*}{Baseline} 
 & 8            & 63.9 & 75.4 &3.2   &4.0  \\
 & 16          & 40.8 & 55.7 &8.5     &11.5  \\
\midrule

Baseline $+$ TP & 4  & 71.8 & 79.7 &87.2   & 88.5 \\
Baseline $+$ NFTP & 4 & 74.2 & 82.2 &92.4   &93.7 \\
\midrule

\multirow{3}{*}{SSR-C} & 4  & 75.5 & 83.9 & 93.5 & 94.3  \\
 & 8 & \textbf{78.7} & \textbf{85.6} & \textbf{95.3} & \textbf{95.9}  \\
 & 16  & 70.9 & 80.3 & 94.9 & 95.3 \\ 
\bottomrule
\end{tabular}}
\label{sampleamount}
\end{table}

\subsubsection{Influence of the Amount of Sampled Frames}
In Tabs.~\ref{moduelablation} and \ref{nf_tp_ablation}, we notice a remarkable performance improvement on the Duke-V dataset after applying NFTP (especially its tracklet partitioning design), while the baseline based on the original tracklets fails to converge. We further investigate the influence of the number of sampled frames and NFTP on the Duke-V dataset in Tab.~\ref{sampleamount}.
Firstly, we try different numbers of sampled frames for each tracklet using the baseline method without applying NFTP. The results show that simply increasing the number of sampled frames for each tracklet is useless for enabling model convergence. 
However, when incorporating tracklet partition, which alleviates the negative impact of unexpected variations within tracklets, a huge performance improvement is observed. Note that on Duke-V, sampling 16 frames per tracklet (Line 2) is roughly equal to the data volume of sampling 4 frames per tracklet after partition (Line 3), but the performance of the former is still unsatisfactory. The results emphasize the indispensable role of the partition design, which alleviates the tracklet feature bias by dividing the tracklets into sub-tracklets. 
Additionally, the NFTP module better promotes learning with noisy frames being filtered, as shown in Line 4.

We also investigate the influence of the number of sampled frames on the MARS dataset, where the intra-tracklet variations are relatively small. For the baseline method, setting the number of sampled frames per tracklet large will lead to degraded performance. Excessive sampling would introduce redundant frames for short tracklets, and unexpected variations and noises within tracklets for long tracklets.
When introducing TP, we observe a notable improvement in performance. TP helps alleviate the bias in tracklet-level feature representation, especially when different segments of a tracklet contain varying levels of useful information. Further improvements are achieved by incorporating the noise filtering design, which refines the sub-tracklet representations.

We further explore the influence of the different numbers of sampled frames after incorporating NFTP in our SSR-C framework in Tab.~\ref{sampleamount}. Note that after applying NFTP, we sample frames in the sub-tracklet level, rather than the original tracklet. 
Sampling 8 frames per sub-tracklet yields the best performance. This configuration strikes a good balance between data diversity and training efficiency. The results show the significant performance gains of our NFTP design, demonstrating its effectiveness. By adaptively aggregating noise-filtered sub-tracklets, our proposed method unleashes the potential of tracklets and promotes unsupervised video person re-identification effectively.
Our proposed method does not have any assumption on the dataset property. It achieves good results on both MARS and Duke-V, although the two datasets have different tracklet distributions.

\subsubsection{Influence of Different Pooling Strategies}
GeM pooling generalizes max and average pooling and is more robust to noises in the unsupervised scenario, so it is adopted in SSR-C to obtain the feature embedding of each frame. We compare different pooling strategies in Tab.~\ref{pooling_abalation} and find GeM more effective. Note that by adopting either GAP or GMP, our method still surpasses state-of-the-art methods on both datasets by a large margin, showing its effectiveness and robustness.

\begin{table}[t]
\centering
\caption{\textbf{Influence of different pooling strategies.} GAP and GMP denote global average and max pooling, respectively. 
}
\setlength{\tabcolsep}{4.8mm}{
\begin{tabular}{lcccc}
\toprule
\multirow{2}{*}{\textbf{Poolings}} & \multicolumn{2}{c}{\textbf{MARS}} & \multicolumn{2}{c}{\textbf{Duke-V}} \\ \cmidrule(r){2-3} \cmidrule(r){4-5} & mAP   & Rank-1  & mAP    & Rank-1  \\ 
\midrule

GAP            & 72.4 & 80.5 & 94.5 & 95.2 \\
GMP            & 77.1 & 83.3 & 93.7 & 94.2 \\
GeM & \textbf{78.7} & \textbf{85.6} & \textbf{95.3} & \textbf{95.9}\\ \bottomrule

\end{tabular}}
\label{pooling_abalation}
\end{table}

\begin{figure*}[t]
  \centering
  \includegraphics[width=0.85\linewidth]{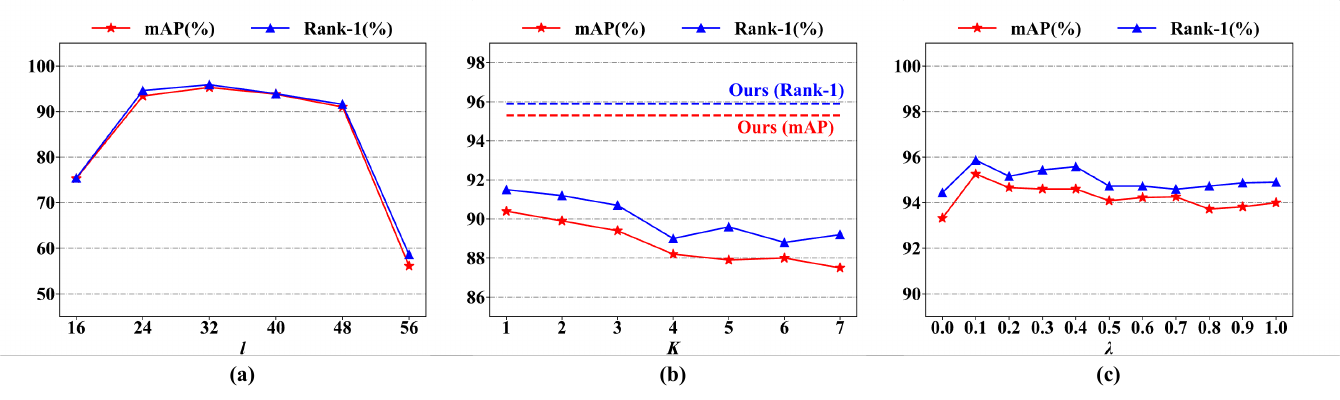}
  \caption{\textbf{Influence of different values of (a) $l$, (b) $K$, (c) $\lambda$.} The blue and red dashed lines in (b) denote the Rank-1 and mAP accuracies of our proposed method, respectively. The ablation experiments are performed on the Duke-V dataset and the optimal hyper-parameter values are directly applied to MARS.}
  \label{fig:3param}
\end{figure*}

\subsubsection{Analysis of $l$}
Tracklet partitioning stride $l$ controls the granularity of sub-tracklet representation. As shown in Fig.~\ref{fig:3param}~(a), large $l$ exacerbates variations within tracklets and contributes to inferior performance, which confirms our motivation. Conversely, small $l$ brings redundancy and a massive data column. For a tradeoff between performance and efficiency, we set $l=32$.

\subsubsection{Analysis of $K$}
The value of $K$ in our CSC loss is dynamically determined considering different numbers of positive sub-tracklets for different people. Besides dynamicity, CSC loss guides learning at the fine-grained sub-tracklet level. We compare with methods using a fixed $K$ design~\cite{li2019unsupervised,lin2020unsupervised} in Tab.~\ref{comparsion} and SSR-C shows great superiority.
We also try manually setting $K$ with different values in Fig.~\ref{fig:3param}~(b). They all perform worse and the optimal $K$ needs cherry-pick.

\subsubsection{Analysis of $\lambda$}
As stated in Eq. \ref{eq:labeldistribution}, $\lambda$ controls the weight of smoothing assigned to $K$ positive sub-clusters. When $\lambda=0$, the model does not consider other positive sub-tracklets from the same tracklet or different tracklets with the same identity. Since it pushes different sub-clusters with the same identity away, it achieves sub-optimal results as shown in Fig.~\ref{fig:3param}~(c). Compared to the widely used InfoNCE loss, CSC loss shows better performance for different choices of $\lambda$.
As the value of $\lambda$ increases, higher weights are assigned to other potential positive sub-clusters, which would distract the model learning, so we observe a downward trend in performance in Fig. \ref{fig:3param}~(c). The best results are achieved when $\lambda=0.1$.

\begin{figure}[t]
\centering
  \includegraphics[width=1\linewidth]{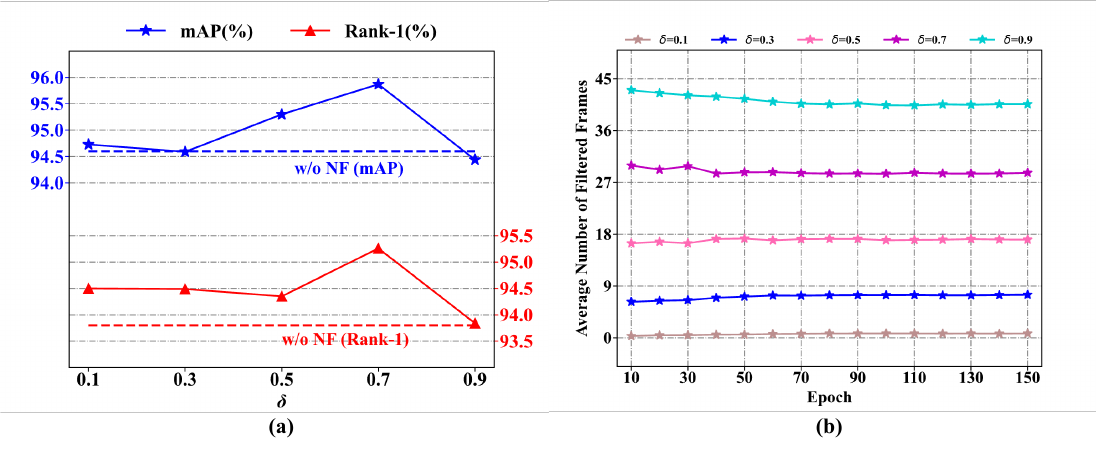}
  \caption{\textbf{(a) Influence of different values of $\delta$ on the performance, and (b) the average number of filtered frames at different epochs with various values of $\delta$}. The blue and red dashed lines in (a) denote the Rank-1 and mAP accuracies of the variant without using NF, respectively. The ablation is performed on Duke-V and the optimal value are directly applied to MARS without further tuning.}
  \label{fig:2param}
\end{figure}

\subsubsection{Analysis of $\delta$}
\label{subsubsec:analyse_delta}
We try various values of $\delta$ for noise filtering in Fig.~\ref{fig:2param}~(a), and note an overall performance improvement when applying NF. According to Eq.~\ref{eq:filteringthreshold}, as the value of $\delta$ increases, more frames would be identified as noises and filtered.
When $\delta$ is relatively small, some noisy frames cannot be filtered, which would misguide clustering and result in sub-optimal results. As $\delta$ is overly large, the model also achieves inferior results, since more frames are identified as noises and some frames that contain useful information may be filtered excessively.
In our experiments, noises within tracklets can be effectively filtered with $\delta = 0.7$.

\subsubsection{Amount of Filtered Noisy Frames}
The results in Fig.~\ref{fig:2param} (b) support our deduction that more frames can be filtered as $\delta$ increases. We also observe that even when $\delta$ remains constant, the number of filtered noisy frames varies dynamically as the training progresses.
When $\delta$ is less than 0.5, more noisy frames are filtered in the later training stage. However, when the value of $\delta$ is larger than 0.5, the number of filtered noisy frames gradually decreases along with the training process.
Intuitively, relatively easier frames should be used for model learning at first, and as the feature extraction capability of the model improves, some moderately hard frames can be incorporated to enhance the robustness of the model. When $\delta=0.7$, which is the value we use, the trend of the number of filtered noisy frames matches the above expectation.

\begin{figure}[t]
\centering
  \includegraphics[width=1\linewidth]{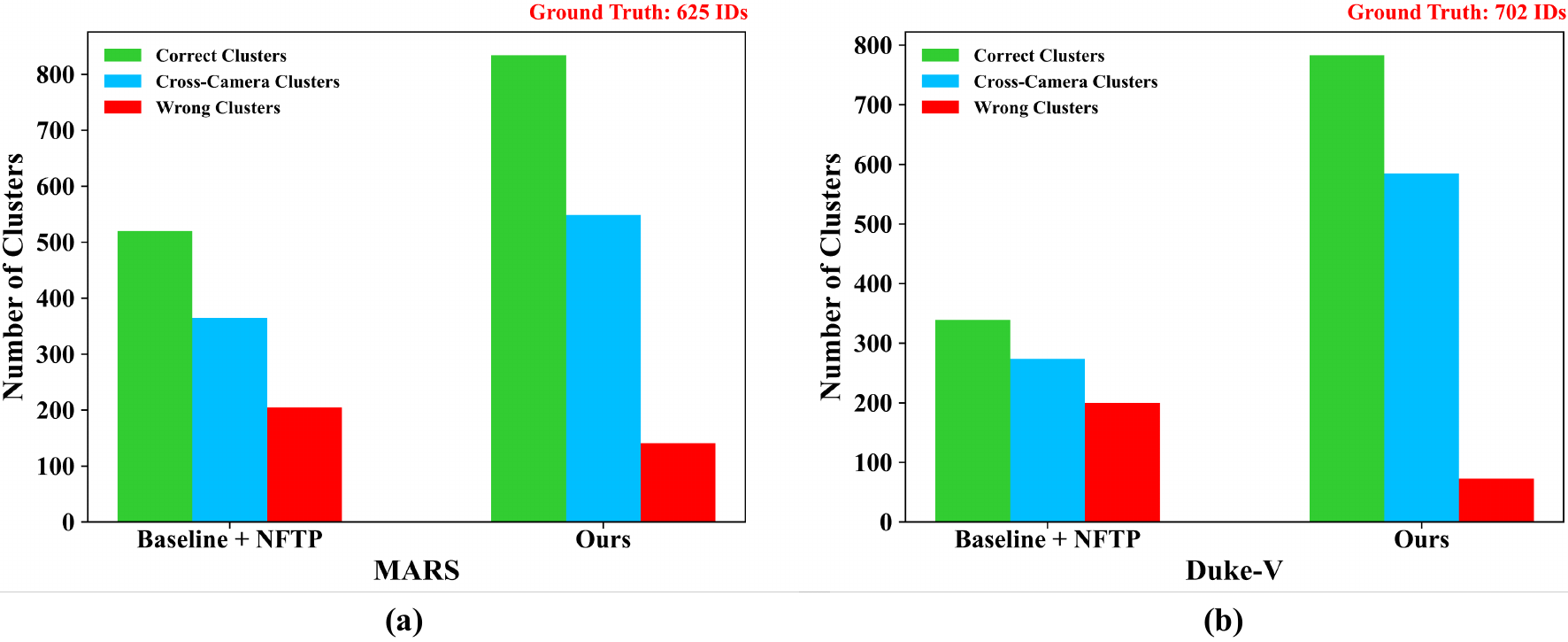}
  \caption{\textbf{The statistical information of clustering.} The number of correct clusters, cross-camera clusters, and incorrect clusters using ``Baseline $+$ NFTP'' and our method are displayed, respectively. For correct clusters, all the samples within the cluster share the same identity label. Cross-camera clusters are a subset of correct clusters. For each dataset, the ground truth number of identities is shown in the upper right.}
  \label{fig:statistic}
\end{figure}

\subsubsection{Statistics of Clustering}
To intuitively show why SSR-C can better guide model learning, we also compare its statistics of clustering with ``Baseline $+$ NFTP'' (Since without NFTP, the model would not converge on Duke-V as discussed in Sec.~\ref{subsec:ablation}). As shown in Fig.~\ref{fig:statistic}, SSR-C contributes to a larger proportion of correct clusters on both datasets. It also exhibits significantly fewer incorrect clusters, which demonstrates the efficacy of SSR-C in improving clustering results.
We also compare the number of clusters with cross-camera samples, \textit{i.e.}, cross-camera clusters. The more correct cross-camera clusters of SSR-C can help our model learn richer information from multiple camera views and promote cross-camera identity matching. 

\begin{table}[t]
\centering
\caption{\textbf{Ablation studies of different clustering algorithms.} The experiments are conducted on the Duke-V dataset. ``\#Clusters'' denotes the preset number of clusters.}
\setlength{\tabcolsep}{4.5mm}{
    \begin{tabular}{ccccc}
    \toprule
    \textbf{Clustering Alg.} & \textbf{\#Clusters} & \textbf{mAP} & \textbf{Rank-1} \\ \midrule
    
    \multirow{8}{*}{K-means} & 500 & 7.8 & 11.8 \\
     & 600 & 21.5 & 26.2 \\
     & 700 & 93.0 & 93.7 \\
     & 800 & 93.0 & 94.2 \\
     & 900 & 93.3 & 93.3 \\
     & 1000 & 94.7 & 94.9 \\
     & 1100 & 93.3 & 93.9 \\
     & 1200 & 91.5 & 92.9 \\
    \midrule
    
    DBSCAN & Dynamic & \textbf{95.3}  & \textbf{95.9}  \\ \bottomrule
    \end{tabular}}
\label{tab:ablation_kmeans}
\end{table}

\subsubsection{Ablation of the Clustering Algorithm}
To show the effectiveness of applying DBSCAN~\cite{ester1996density} for clustering in our proposed SSR-C framework, we also try replacing it with K-means~\cite{macqueen1967some}, which needs presetting the number of clusters. As shown in Tab.~\ref{tab:ablation_kmeans}, when the number of clusters is too small, different identities are incorrectly grouped together, weakening the model's ability to distinguish between individuals. 
While increasing the number of clusters can mitigate this to some extent, the fixed nature of K-means clustering causes it to push away features from the same identity in the whole training process, harming discriminative learning. 
Additionally, searching for the optimal number of clusters requires prior information of the data and many preliminary experiments, making K-means inflexible and unrealistic in applications. As shown in Tab.~\ref{tab:ablation_kmeans}, when the number of clusters is not appropriately set (\textit{e.g.}, 500 and 600), the performance degrades significantly. 
In contrast, DBSCAN dynamically adjusts the number of clusters based on the underlying feature distribution, allowing for more accurate and adaptive clustering. Combined with our proposed self-supervised refined clustering design, this flexibility enables the model to continuously refine clustering results in response to improved feature representations, leading to better performance in the unsupervised setting.
The results show that our SSR-C based on DBSCAN is capable of generating reliable pseudo labels to achieve effective feature learning.

\subsection{Visualizations}
In this section, to intuitively demonstrate the superiority of our proposed method, we provide visualizations of noisy frames, clustering and retrieval results on MARS, along with in-depth analysis and discussions.

\begin{figure}[t]
\centering
  \includegraphics[width=0.8\linewidth]{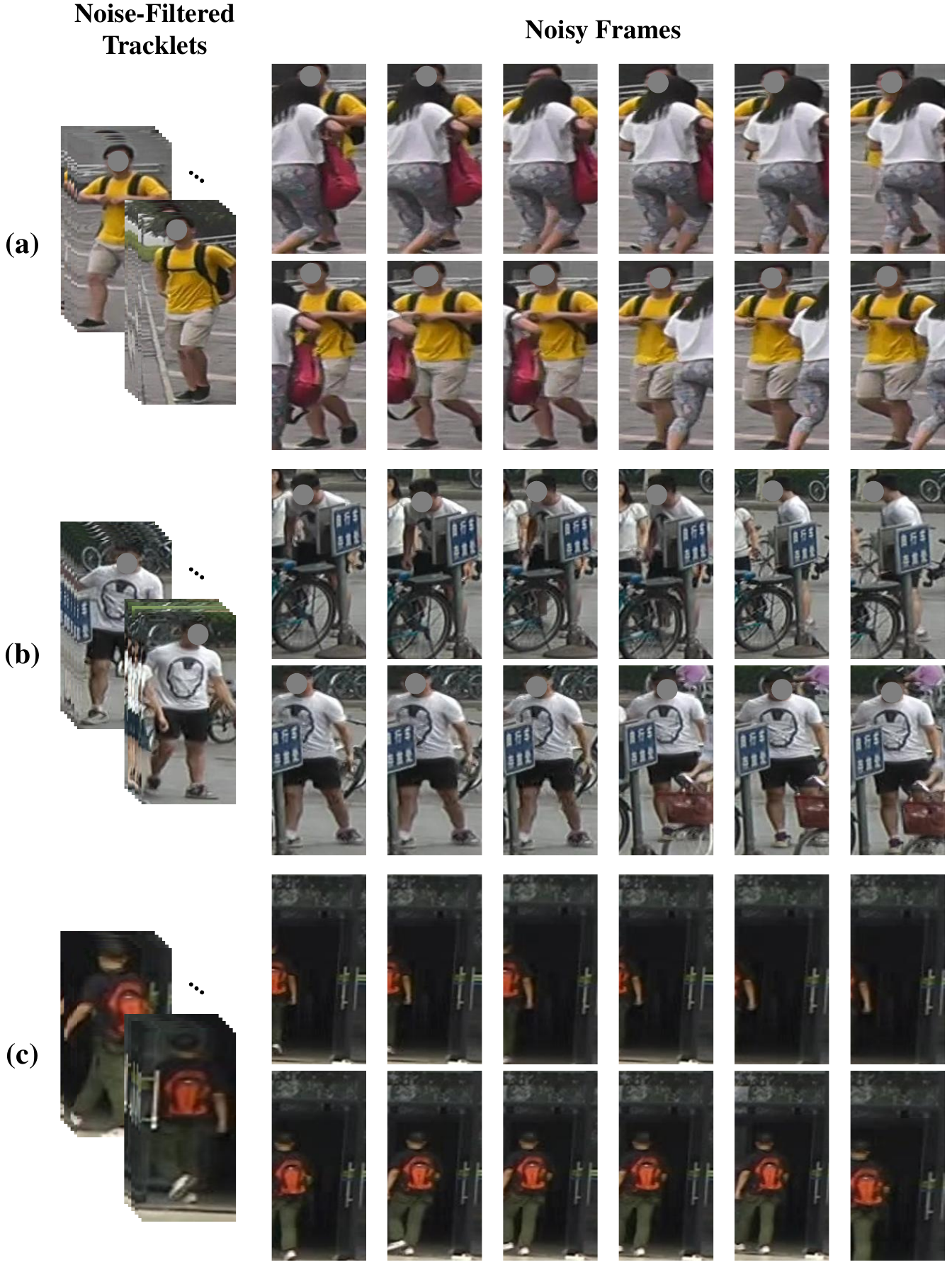}
  \caption{\textbf{Visualizations of filtered noisy frames.}
  For each tracklet, the first row displays the heavily noisy frames that are filtered during the whole training stage, and the second row displays some frames with relatively mild noise that are only filtered during the early stage.}
  \label{fig:noise-filter}
  \vspace{-0.05in}
\end{figure}

\subsubsection{Visualization of Noisy Frames}
The presence of noisy frames can render each tracklet feature non-discriminative, misleading the clustering algorithm to generate incorrect pseudo labels and disturbing feature learning. Especially when we apply our proposed tracklet partition, sub-tracklets with noises may cause relay errors in the following sub-cluster merging step.
Therefore, noise filtering plays a pivotal role in SSR-C. To intuitively understand how it works and show its effectiveness, we visualize the filtered noisy frames at different training stages in Fig.~\ref{fig:noise-filter}.
As shown in Fig.~\ref{fig:noise-filter} (a) and (b), there may be occlusions caused by other pedestrians or objects. If we perform clustering based on the unfiltered tracklet features, tracklets belonging to different individuals may be close in the embedding space and be wrongly clustered. Similarly, frames with great scale changes and background interference can also be regarded as noises and be filtered, as shown in Fig.~\ref{fig:noise-filter} (c).

Additionally, some slightly noisy frames are discarded during the early training stages, but are included for training later. For example, as shown in the first row of Fig.~\ref{fig:noise-filter} (a) and (b), there are heavy occlusions from other pedestrians or objects, thereby only limited identity information remains in them. Since there is no supervision of ground-truth identities, these noisy frames can interfere with the tracklet features, which in turn interferes with reliable clustering and disrupts the model optimization. 
Nevertheless, as shown in the second row of Fig.~\ref{fig:noise-filter} (a) and (b), some noisy frames can still contain useful identity information. With moderate noises, they can help the model deal with occlusion interference in the real world. Therefore, thanks to our dynamic noise-filtering design, they can be used for robust feature learning. 
Another example is shown in Fig.~\ref{fig:noise-filter} (c), where with the pedestrian walking into the room and away from the camera, the scale varies and the background interference increases. As the training progresses, some difficult noisy frames (in the first row) are filtered all the way, while some moderately challenging frames (in the second row) are gradually used for subsequent model optimization.
Since the model's capability and robustness improve as training, gradually introducing relatively hard frames can improve its robustness in dealing with various interference.

\begin{figure}[t]
\centering
  \includegraphics[width=1\linewidth]{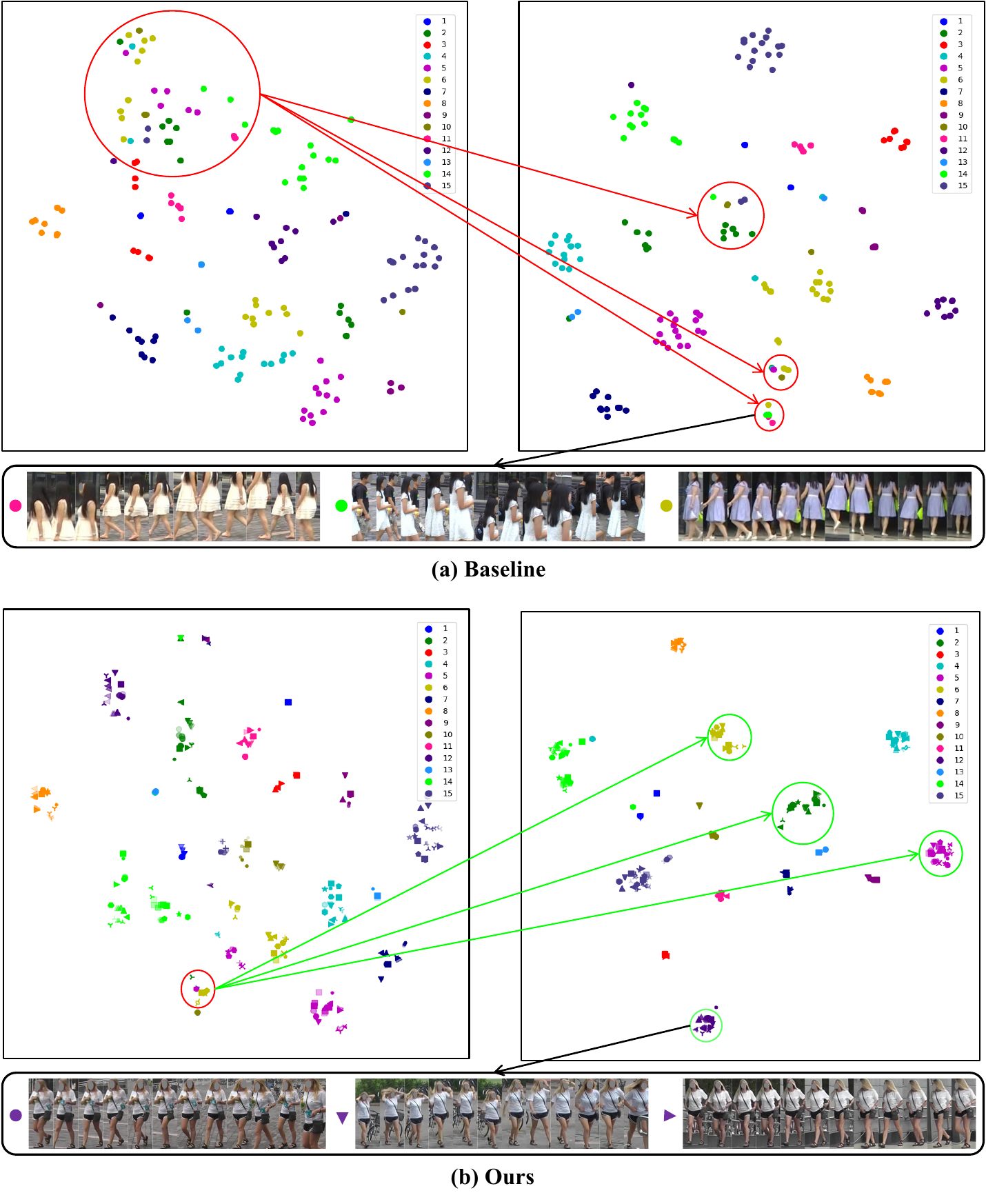}
  \caption{\textbf{Clustering results of (a) the baseline and (b) our proposed SSR-C.}
  Different colors denote different identities and different shapes represent different tracklets. Various transparencies of the same color and shape denote the sub-tracklets of each tracklet. The green/red circles show the correct/wrong results. The left/right results of each sub-figure correspond to the early/late training stage. Best viewed in color and zoomed in.
  }
  \label{fig:cluster}
\end{figure}

\subsubsection{Clustering Results}
We also use t-SNE~\cite{van2008visualizing} to visualize the clustering results in Fig.~\ref{fig:cluster}. For the baseline based on tracklets, since there are noises within tracklets, different identities are aggregated together incorrectly at the early stage in Fig.~\ref{fig:cluster}~(a). When it comes to the later training stage, the clustering results are still unsatisfactory with a lot of wrong clusters. 
In contrast, for SSR-C, multiple sub-tracklets from the same tracklet, acting as intra-class cross-tracklet relays, can effectively aggregate various cross-camera tracklets of the same identity as shown at the bottom of Fig.~\ref{fig:cluster}~(b). Additionally, although some tracklets with different identities are wrongly clustered in the early training stage, they are separated in the later training stage. 
By comparing the clustering results at the late training stage in Fig.~\ref{fig:cluster}, we can observe that SSR-C contributes to better intra-class compactness and inter-class separability. Therefore, SSR-C can generate more accurate pseudo labels and promote the feature extraction capability of the model.

\begin{figure}[t]
\centering
  \includegraphics[width=1\linewidth]{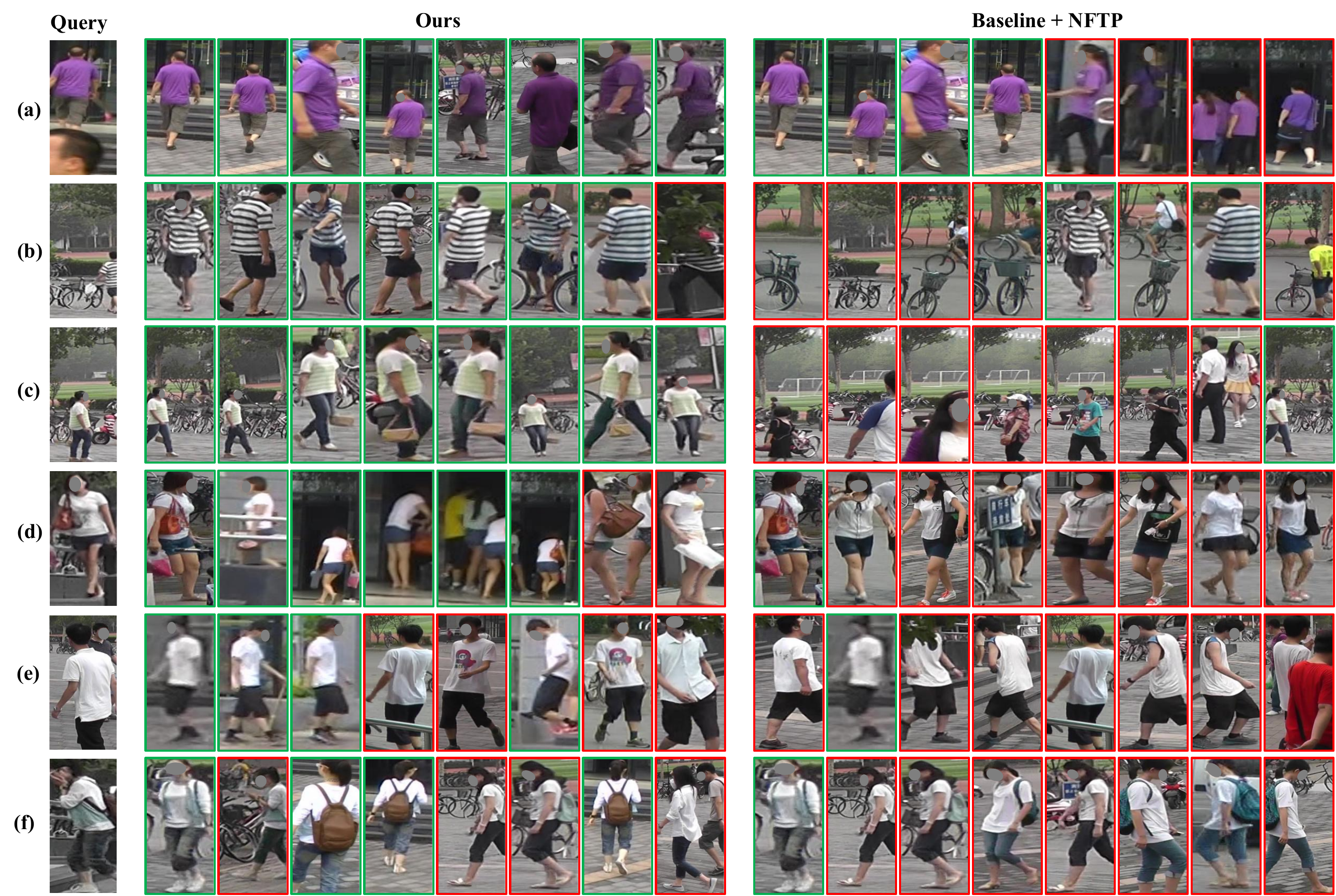}
  \caption{\textbf{Visualization of top eight retrieval results of our proposed SSR-C and ``Baseline + NFTP'' on MARS, respectively.} The correct and wrong results are in green and red boxes, respectively. To simplify the presentation, we showcase only the first frame of each tracklet. Best viewed in color and zoomed in.}
  \label{fig:rank}
\end{figure}

\subsubsection{Retrieval Results}
To intuitively demonstrate the effectiveness of our method, we visualize the retrieval results in Fig.~\ref{fig:rank}. Considering the poor performance of the baseline, we compare the retrieval results between our proposed SSR-C and ``Baseline $+$ NFTP''. 
As shown in Fig.~\ref{fig:rank} (a), SSR-C can better encourage the model to learn useful identity information, while ``Baseline $+$ NFTP'' may be influenced by similar color appearance information. The results of (b) and (c) show that our SSR-C is more capable of attending to the effective regions of pedestrians, and the results of (c) and (d) demonstrate its effectiveness in handling view changes and scene variations. 
Due to poor clustering accuracy, ``Baseline $+$ NFTP'' cannot handle complicated variations and visual interference, leading to wrong matches. 
Overall, the results validate the heightened robustness of our proposed SSR-C across diverse challenging scenarios, including lighting variations, scene changes, viewpoint changes, similar appearances, and misalignment induced by tracking. 
We also observe some failure cases of SSR-C in (e) and (f), where people with similar body shapes wear similar clothes or carry similar backpacks. It is expected to improve its robustness by incorporating other auxiliary information such as attributes and part-based modules. 
Without relying on extra information, SSR-C demonstrates improved feature learning and retrieval results.

\section{Conclusion and Outlook}
To overcome the dependence on identity labels and exploit rich information naturally in tracklets, we propose the Self-Supervised Refined Clustering (SSR-C) framework for unsupervised video person Re-ID. It unleashes the potential of tracklets for effective model optimization.
Specifically, we first design the NFTP module to eliminate the noises in tracklets and partition each tracklet into multiple sub-tracklets. With noise-filtered sub-tracklets, we take advantage of the self-supervised signal that the sub-tracklets from the same tracklet indicate a consistent identity, and propose a progressive sub-cluster merging strategy to achieve reliable clustering and generate accurate pseudo labels. To exploit the associations between sub-tracklets and facilitate feature learning, we further propose CSC loss. 
Extensive experiments demonstrate the effectiveness and superiority of our proposed SSR-C. Without relying on manual annotations or auxiliary information, SSR-C approaches advanced supervised methods in performance. There is no additional overhead during inference, making it efficient in real-world applications. 

In our SSR-C framework, we propose a straightforward strategy for tracklet partition, \textit{i.e.}, sequentially dividing each noise-filtered tracklet into multiple sub-tracklets. It would be promising to explore alternative partitioning strategies in future work, to better handle the variations present in long tracklets.
Moreover, as illustrated in Figs.~\ref{fig:2param}~(b) and \ref{fig:noise-filter}, the number of filtered noisy frames varies throughout the training process. Although our noise filtering design relies solely on tracklet features, a promising direction for future work is to dynamically filter noisy frames at different stages of training.

\bibliographystyle{IEEEtran}
\bibliography{bib}

\begin{thebibliography}{10}
\providecommand{\url}[1]{#1}
\csname url@samestyle\endcsname
\providecommand{\newblock}{\relax}
\providecommand{\bibinfo}[2]{#2}
\providecommand{\BIBentrySTDinterwordspacing}{\spaceskip=0pt\relax}
\providecommand{\BIBentryALTinterwordstretchfactor}{4}
\providecommand{\BIBentryALTinterwordspacing}{\spaceskip=\fontdimen2\font plus
\BIBentryALTinterwordstretchfactor\fontdimen3\font minus \fontdimen4\font\relax}
\providecommand{\BIBforeignlanguage}[2]{{%
\expandafter\ifx\csname l@#1\endcsname\relax
\typeout{** WARNING: IEEEtran.bst: No hyphenation pattern has been}%
\typeout{** loaded for the language `#1'. Using the pattern for}%
\typeout{** the default language instead.}%
\else
\language=\csname l@#1\endcsname
\fi
#2}}
\providecommand{\BIBdecl}{\relax}
\BIBdecl

\bibitem{zheng2016person}
L.~Zheng, Y.~Yang, and A.~G. Hauptmann, ``Person re-identification: Past, present and future,'' \emph{arXiv preprint arXiv:1610.02984}, 2016.

\bibitem{ye2021deep}
M.~Ye, J.~Shen, G.~Lin, T.~Xiang, L.~Shao, and S.~C. Hoi, ``Deep learning for person re-identification: A survey and outlook,'' \emph{IEEE Transactions on Pattern Analysis and Machine Intelligence}, vol.~44, no.~6, pp. 2872--2893, 2021.

\bibitem{qian2019leader}
X.~Qian, Y.~Fu, T.~Xiang, Y.-G. Jiang, and X.~Xue, ``Leader-based multi-scale attention deep architecture for person re-identification,'' \emph{IEEE Transactions on Pattern Analysis and Machine Intelligence}, vol.~42, no.~2, pp. 371--385, 2019.

\bibitem{he2021transreid}
S.~He, H.~Luo, P.~Wang, F.~Wang, H.~Li, and W.~Jiang, ``Transreid: Transformer-based object re-identification,'' in \emph{Proceedings of the IEEE/CVF International Conference on Computer Vision}, 2021, pp. 15\,013--15\,022.

\bibitem{wang2022co}
Q.~Wang, X.~Qian, Y.~Fu, and X.~Xue, ``Co-attention aligned mutual cross-attention for cloth-changing person re-identification,'' in \emph{Proceedings of the Asian Conference on Computer Vision}, 2022, pp. 2270--2288.

\bibitem{wang2024exploring}
Q.~Wang, X.~Qian, B.~Li, X.~Xue, and Y.~Fu, ``Exploring fine-grained representation and recomposition for cloth-changing person re-identification,'' \emph{IEEE Transactions on Information Forensics and Security}, vol.~19, pp. 6280--6292, 2024.

\bibitem{wang2025content}
Q.~Wang, X.~Qian, B.~Li, L.~Chen, Y.~Fu, and X.~Xue, ``Content and salient semantics collaboration for cloth-changing person re-identification,'' in \emph{IEEE International Conference on Acoustics, Speech and Signal Processing}.\hskip 1em plus 0.5em minus 0.4em\relax IEEE, 2025, pp. 1--5.

\bibitem{wang2025distribution}
Q.~Wang, X.~Qian, B.~Li, and X.~Xue, ``Distribution aligned semantics adaption for lifelong person re-identification,'' \emph{Machine Learning}, vol. 114, no.~3, pp. 1--22, 2025.

\bibitem{wang2025image}
Q.~Wang, X.~Qian, B.~Li, Y.~Fu, and X.~Xue, ``Image-text-image knowledge transfer for lifelong person re-identification with hybrid clothing states,'' \emph{IEEE Transactions on Image Processing}, vol.~34, pp. 5584--5597, 2025.

\bibitem{zhao2016person}
R.~Zhao, W.~Oyang, and X.~Wang, ``Person re-identification by saliency learning,'' \emph{IEEE Transactions on Pattern Analysis and Machine Intelligence}, vol.~39, no.~2, pp. 356--370, 2016.

\bibitem{sun2018beyond}
Y.~Sun, L.~Zheng, Y.~Yang, Q.~Tian, and S.~Wang, ``Beyond part models: Person retrieval with refined part pooling (and a strong convolutional baseline),'' in \emph{Proceedings of the European Conference on Computer Vision}, 2018, pp. 480--496.

\bibitem{zhou2019omni}
K.~Zhou, Y.~Yang, A.~Cavallaro, and T.~Xiang, ``Omni-scale feature learning for person re-identification,'' in \emph{Proceedings of the IEEE/CVF International Conference on Computer Vision}, 2019, pp. 3702--3712.

\bibitem{shu2021large}
X.~Shu, X.~Wang, X.~Zang, S.~Zhang, Y.~Chen, G.~Li, and Q.~Tian, ``Large-scale spatio-temporal person re-identification: Algorithms and benchmark,'' \emph{IEEE Transactions on Circuits and Systems for Video Technology}, vol.~32, no.~7, pp. 4390--4403, 2021.

\bibitem{wang2023rethinking}
Q.~Wang, X.~Qian, B.~Li, Y.~Fu, and X.~Xue, ``Rethinking person re-identification from a projection-on-prototypes perspective,'' \emph{arXiv preprint arXiv:2308.10717}, 2023.

\bibitem{yan2020learning}
Y.~Yan, J.~Qin, J.~Chen, L.~Liu, F.~Zhu, Y.~Tai, and L.~Shao, ``Learning multi-granular hypergraphs for video-based person re-identification,'' in \emph{Proceedings of the IEEE/CVF Conference on Computer Vision and Pattern Recognition}, 2020, pp. 2899--2908.

\bibitem{yang2020spatial}
J.~Yang, W.-S. Zheng, Q.~Yang, Y.-C. Chen, and Q.~Tian, ``Spatial-temporal graph convolutional network for video-based person re-identification,'' in \emph{Proceedings of the IEEE/CVF Conference on Computer Vision and Pattern Recognition}, 2020, pp. 3289--3299.

\bibitem{wu2020adaptive}
Y.~Wu, O.~E.~F. Bourahla, X.~Li, F.~Wu, Q.~Tian, and X.~Zhou, ``Adaptive graph representation learning for video person re-identification,'' \emph{IEEE Transactions on Image Processing}, vol.~29, pp. 8821--8830, 2020.

\bibitem{hou2021bicnet}
R.~Hou, H.~Chang, B.~Ma, R.~Huang, and S.~Shan, ``Bicnet-tks: Learning efficient spatial-temporal representation for video person re-identification,'' in \emph{Proceedings of the IEEE/CVF Conference on Computer Vision and Pattern Recognition}, 2021, pp. 2014--2023.

\bibitem{zang2022exploiting}
X.~Zang, G.~Li, W.~Gao, and X.~Shu, ``Exploiting robust unsupervised video person re-identification,'' \emph{IET Image Processing}, vol.~16, no.~3, pp. 729--741, 2022.

\bibitem{hou2019vrstc}
R.~Hou, B.~Ma, H.~Chang, X.~Gu, S.~Shan, and X.~Chen, ``Vrstc: Occlusion-free video person re-identification,'' in \emph{Proceedings of the IEEE/CVF Conference on Computer Vision and Pattern Recognition}, 2019, pp. 7183--7192.

\bibitem{wu2019progressive}
Y.~Wu, Y.~Lin, X.~Dong, Y.~Yan, W.~Bian, and Y.~Yang, ``Progressive learning for person re-identification with one example,'' \emph{IEEE Transactions on Image Processing}, vol.~28, no.~6, pp. 2872--2881, 2019.

\bibitem{zeng2022anchor}
S.~Zeng, X.~Wang, M.~Liu, Q.~Liu, and Y.~Wang, ``Anchor association learning for unsupervised video person re-identification,'' \emph{IEEE Transactions on Neural Networks and Learning Systems}, vol.~35, no.~1, pp. 1013--1024, 2022.

\bibitem{wu2018exploit}
Y.~Wu, Y.~Lin, X.~Dong, Y.~Yan, W.~Ouyang, and Y.~Yang, ``Exploit the unknown gradually: One-shot video-based person re-identification by stepwise learning,'' in \emph{Proceedings of the IEEE Conference on Computer Vision and Pattern Recognition}, 2018, pp. 5177--5186.

\bibitem{ye2017dynamic}
M.~Ye, A.~J. Ma, L.~Zheng, J.~Li, and P.~C. Yuen, ``Dynamic label graph matching for unsupervised video re-identification,'' in \emph{Proceedings of the IEEE International Conference on Computer Vision}, 2017, pp. 5142--5150.

\bibitem{liu2017stepwise}
Z.~Liu, D.~Wang, and H.~Lu, ``Stepwise metric promotion for unsupervised video person re-identification,'' in \emph{Proceedings of the IEEE International Conference on Computer Vision}, 2017, pp. 2429--2438.

\bibitem{ye2018robust}
M.~Ye, X.~Lan, and P.~C. Yuen, ``Robust anchor embedding for unsupervised video person re-identification in the wild,'' in \emph{Proceedings of the European Conference on Computer Vision}, 2018, pp. 170--186.

\bibitem{li2019unsupervised}
M.~Li, X.~Zhu, and S.~Gong, ``Unsupervised tracklet person re-identification,'' \emph{IEEE Transactions on Pattern Analysis and Machine Intelligence}, vol.~42, no.~7, pp. 1770--1782, 2019.

\bibitem{chen2018deep}
Y.~Chen, X.~Zhu, and S.~Gong, ``Deep association learning for unsupervised video person re-identification,'' \emph{arXiv preprint arXiv:1808.07301}, 2018.

\bibitem{li2022cluster}
M.~Li, C.-G. Li, and J.~Guo, ``Cluster-guided asymmetric contrastive learning for unsupervised person re-identification,'' \emph{IEEE Transactions on Image Processing}, vol.~31, pp. 3606--3617, 2022.

\bibitem{ge2020self}
Y.~Ge, F.~Zhu, D.~Chen, R.~Zhao \emph{et~al.}, ``Self-paced contrastive learning with hybrid memory for domain adaptive object re-id,'' \emph{Advances in Neural Information Processing Systems}, vol.~33, pp. 11\,309--11\,321, 2020.

\bibitem{cho2022part}
Y.~Cho, W.~J. Kim, S.~Hong, and S.-E. Yoon, ``Part-based pseudo label refinement for unsupervised person re-identification,'' in \emph{Proceedings of the IEEE/CVF Conference on Computer Vision and Pattern Recognition}, 2022, pp. 7308--7318.

\bibitem{chen2021ice}
H.~Chen, B.~Lagadec, and F.~Bremond, ``Ice: Inter-instance contrastive encoding for unsupervised person re-identification,'' in \emph{Proceedings of the IEEE/CVF International Conference on Computer Vision}, 2021, pp. 14\,960--14\,969.

\bibitem{lan2023learning}
L.~Lan, X.~Teng, J.~Zhang, X.~Zhang, and D.~Tao, ``Learning to purification for unsupervised person re-identification,'' \emph{IEEE Transactions on Image Processing}, vol.~32, pp. 3338--3353, 2023.

\bibitem{fan2018unsupervised}
H.~Fan, L.~Zheng, C.~Yan, and Y.~Yang, ``Unsupervised person re-identification: Clustering and fine-tuning,'' \emph{ACM Transactions on Multimedia Computing, Communications, and Applications}, vol.~14, no.~4, pp. 1--18, 2018.

\bibitem{lin2019bottom}
Y.~Lin, X.~Dong, L.~Zheng, Y.~Yan, and Y.~Yang, ``A bottom-up clustering approach to unsupervised person re-identification,'' in \emph{Proceedings of the AAAI Conference on Artificial Intelligence}, vol.~33, no.~01, 2019, pp. 8738--8745.

\bibitem{wu2020tracklet}
G.~Wu, X.~Zhu, and S.~Gong, ``Tracklet self-supervised learning for unsupervised person re-identification,'' in \emph{Proceedings of the AAAI Conference on Artificial Intelligence}, vol.~34, no.~07, 2020, pp. 12\,362--12\,369.

\bibitem{xie2021unsupervised}
P.~Xie, X.~Xu, Z.~Wang, and T.~Yamasaki, ``Unsupervised video person re-identification via noise and hard frame aware clustering,'' in \emph{Proceedings of the IEEE International Conference on Multimedia and Expo}.\hskip 1em plus 0.5em minus 0.4em\relax IEEE, 2021, pp. 1--6.

\bibitem{ding2019towards}
G.~Ding, S.~Khan, Z.~Tang, J.~Zhang, and F.~Porikli, ``Towards better validity: Dispersion based clustering for unsupervised person re-identification,'' \emph{arXiv preprint arXiv:1906.01308}, 2019.

\bibitem{lin2020unsupervised}
Y.~Lin, L.~Xie, Y.~Wu, C.~Yan, and Q.~Tian, ``Unsupervised person re-identification via softened similarity learning,'' in \emph{Proceedings of the IEEE/CVF Conference on Computer Vision and Pattern Recognition}, 2020, pp. 3390--3399.

\bibitem{dai2022cluster}
Z.~Dai, G.~Wang, W.~Yuan, S.~Zhu, and P.~Tan, ``Cluster contrast for unsupervised person re-identification,'' in \emph{Proceedings of the Asian Conference on Computer Vision}, 2022, pp. 1142--1160.

\bibitem{xie2022sampling}
P.~Xie, X.~Xu, Z.~Wang, and T.~Yamasaki, ``Sampling and re-weighting: Towards diverse frame aware unsupervised video person re-identification,'' \emph{IEEE Transactions on Multimedia}, vol.~24, pp. 4250--4261, 2022.

\bibitem{wang2023relation}
X.~Wang, M.~Liu, F.~Wang, J.~Dai, A.-A. Liu, and Y.~Wang, ``Relation-preserving feature embedding for unsupervised person re-identification,'' \emph{IEEE Transactions on Multimedia}, vol.~26, pp. 714--723, 2023.

\bibitem{macqueen1967some}
J.~MacQueen \emph{et~al.}, ``Some methods for classification and analysis of multivariate observations,'' in \emph{Proceedings of the Fifth Berkeley Symposium on Mathematical Statistics and Probability}, vol.~1, no.~14.\hskip 1em plus 0.5em minus 0.4em\relax Oakland, CA, USA, 1967, pp. 281--297.

\bibitem{tao2024unsupervised}
X.~Tao, J.~Kong, M.~Jiang, M.~Lu, and A.~Mian, ``Unsupervised learning of intrinsic semantics with diffusion model for person re-identification,'' \emph{IEEE Transactions on Image Processing}, vol.~33, pp. 6705--6719, 2024.

\bibitem{ho2020denoising}
J.~Ho, A.~Jain, and P.~Abbeel, ``Denoising diffusion probabilistic models,'' \emph{Advances in Neural Information Processing Systems}, vol.~33, pp. 6840--6851, 2020.

\bibitem{tang2019unsupervised}
H.~Tang, Y.~Zhao, and H.~Lu, ``Unsupervised person re-identification with iterative self-supervised domain adaptation,'' in \emph{Proceedings of the IEEE/CVF Conference on Computer Vision and Pattern Recognition Workshops}, 2019, pp. 1536--1543.

\bibitem{jiang2020self}
K.~Jiang, T.~Zhang, Y.~Zhang, F.~Wu, and Y.~Rui, ``Self-supervised agent learning for unsupervised cross-domain person re-identification,'' \emph{IEEE Transactions on Image Processing}, vol.~29, pp. 8549--8560, 2020.

\bibitem{dou2023identity}
Z.~Dou, Z.~Wang, Y.~Li, and S.~Wang, ``Identity-seeking self-supervised representation learning for generalizable person re-identification,'' in \emph{Proceedings of the IEEE/CVF International Conference on Computer Vision}, 2023, pp. 15\,847--15\,858.

\bibitem{hermans2017defense}
A.~Hermans, L.~Beyer, and B.~Leibe, ``In defense of the triplet loss for person re-identification,'' \emph{arXiv preprint arXiv:1703.07737}, 2017.

\bibitem{hu2021hard}
Z.~Hu, C.~Zhu, and G.~He, ``Hard-sample guided hybrid contrast learning for unsupervised person re-identification,'' in \emph{IEEE International Conference on Network Intelligence and Digital Content}.\hskip 1em plus 0.5em minus 0.4em\relax IEEE, 2021, pp. 91--95.

\bibitem{ester1996density}
M.~Ester, H.-P. Kriegel, J.~Sander, X.~Xu \emph{et~al.}, ``A density-based algorithm for discovering clusters in large spatial databases with noise.'' in \emph{Proceedings of the International Conference on Knowledge Discovery and Data Mining}, vol.~96, no.~34, 1996, pp. 226--231.

\bibitem{oord2018representation}
A.~v.~d. Oord, Y.~Li, and O.~Vinyals, ``Representation learning with contrastive predictive coding,'' \emph{arXiv preprint arXiv:1807.03748}, 2018.

\bibitem{hinton2015distilling}
G.~Hinton, O.~Vinyals, and J.~Dean, ``Distilling the knowledge in a neural network,'' \emph{arXiv preprint arXiv:1503.02531}, 2015.

\bibitem{zhong2019invariance}
Z.~Zhong, L.~Zheng, Z.~Luo, S.~Li, and Y.~Yang, ``Invariance matters: Exemplar memory for domain adaptive person re-identification,'' in \emph{Proceedings of the IEEE/CVF Conference on Computer Vision and Pattern Recognition}, 2019, pp. 598--607.

\bibitem{zhong2020learning}
------, ``Learning to adapt invariance in memory for person re-identification,'' \emph{IEEE Transactions on Pattern Analysis and Machine Intelligence}, vol.~43, no.~8, pp. 2723--2738, 2020.

\bibitem{szegedy2016rethinking}
C.~Szegedy, V.~Vanhoucke, S.~Ioffe, J.~Shlens, and Z.~Wojna, ``Rethinking the inception architecture for computer vision,'' in \emph{Proceedings of the IEEE Conference on Computer Vision and Pattern Recognition}, 2016, pp. 2818--2826.

\bibitem{zheng2016mars}
L.~Zheng, Z.~Bie, Y.~Sun, J.~Wang, C.~Su, S.~Wang, and Q.~Tian, ``Mars: A video benchmark for large-scale person re-identification,'' in \emph{Proceedings of the European Conference on Computer Vision}.\hskip 1em plus 0.5em minus 0.4em\relax Springer, 2016, pp. 868--884.

\bibitem{ristani2016performance}
E.~Ristani, F.~Solera, R.~Zou, R.~Cucchiara, and C.~Tomasi, ``Performance measures and a data set for multi-target, multi-camera tracking,'' in \emph{Proceedings of the European Conference on Computer Vision}.\hskip 1em plus 0.5em minus 0.4em\relax Springer, 2016, pp. 17--35.

\bibitem{zheng2017unlabeled}
Z.~Zheng, L.~Zheng, and Y.~Yang, ``Unlabeled samples generated by gan improve the person re-identification baseline in vitro,'' in \emph{Proceedings of the IEEE International Conference on Computer Vision}, 2017, pp. 3754--3762.

\bibitem{wang2007shape}
X.~Wang, G.~Doretto, T.~Sebastian, J.~Rittscher, and P.~Tu, ``Shape and appearance context modeling,'' in \emph{IEEE 11th International Conference on Computer Vision}, 2007, pp. 1--8.

\bibitem{market1501}
L.~Zheng, L.~Shen, L.~Tian, S.~Wang, J.~Wang, and Q.~Tian, ``Scalable person re-identification: A benchmark,'' in \emph{Proceedings of the IEEE/CVF International Conference on Computer Vision}, 2015, pp. 1116--1124.

\bibitem{zhao2022gait}
Y.~Zhao, X.~Wang, X.~Yu, C.~Liu, and Y.~Gao, ``Gait-assisted video person retrieval,'' \emph{IEEE Transactions on Circuits and Systems for Video Technology}, vol.~33, no.~2, pp. 897--908, 2022.

\bibitem{pan2023multi}
H.~Pan, Y.~Chen, and Z.~He, ``Multi-granularity graph pooling for video-based person re-identification,'' \emph{Neural Networks}, vol. 160, pp. 22--33, 2023.

\bibitem{zhang2023self}
Y.~Zhang, B.~Ma, L.~Liu, X.~Yi, M.~Li, and Y.~Diao, ``Self-paced uncertainty estimation for one-shot person re-identification,'' \emph{Applied Intelligence}, vol.~53, no.~12, pp. 15\,080--15\,094, 2023.

\bibitem{xiao2017joint}
T.~Xiao, S.~Li, B.~Wang, L.~Lin, and X.~Wang, ``Joint detection and identification feature learning for person search,'' in \emph{Proceedings of the IEEE Conference on Computer Vision and Pattern Recognition}, 2017, pp. 3415--3424.

\bibitem{wang2025noise}
J.~Wang, J.~Wen, W.~Ding, C.~Yu, X.~Zhu, and Z.~Wang, ``Noise perception self-supervised learning for unsupervised person re-identification,'' \emph{IEEE Transactions on Emerging Topics in Computational Intelligence}, pp. 1--11, 2025.

\bibitem{zhang2025dual}
C.~Zhang, Y.~Su, N.~Wang, Y.~Lan, T.~Wang, and A.~Li, ``Dual representation modeling and progressive contrastive learning for unsupervised video person re-identification,'' \emph{Neurocomputing}, p. 130467, 2025.

\bibitem{he2016deep}
K.~He, X.~Zhang, S.~Ren, and J.~Sun, ``Deep residual learning for image recognition,'' in \emph{Proceedings of the IEEE Conference on Computer Vision and Pattern Recognition}, 2016, pp. 770--778.

\bibitem{krizhevsky2017imagenet}
A.~Krizhevsky, I.~Sutskever, and G.~E. Hinton, ``Imagenet classification with deep convolutional neural networks,'' \emph{Communications of the ACM}, vol.~60, no.~6, pp. 84--90, 2017.

\bibitem{radenovic2018fine}
F.~Radenovi{\'c}, G.~Tolias, and O.~Chum, ``Fine-tuning cnn image retrieval with no human annotation,'' \emph{IEEE Transactions on Pattern Analysis and Machine Intelligence}, vol.~41, no.~7, pp. 1655--1668, 2018.

\bibitem{zhong2020random}
Z.~Zhong, L.~Zheng, G.~Kang, S.~Li, and Y.~Yang, ``Random erasing data augmentation,'' in \emph{Proceedings of the AAAI Conference on Artificial Intelligence}, vol.~34, no.~07, 2020, pp. 13\,001--13\,008.

\bibitem{zhong2017re}
Z.~Zhong, L.~Zheng, D.~Cao, and S.~Li, ``Re-ranking person re-identification with k-reciprocal encoding,'' in \emph{Proceedings of the IEEE Conference on Computer Vision and Pattern Recognition}, 2017, pp. 1318--1327.

\bibitem{xiong2024inter}
M.~Xiong, K.~Hu, Z.~Lyu, F.~Fang, Z.~Wang, R.~Hu, and K.~Muhammad, ``Inter-camera identity discrimination for unsupervised person re-identification,'' \emph{ACM Transactions on Multimedia Computing, Communications and Applications}, vol.~20, no.~8, pp. 1--18, 2024.

\bibitem{van2008visualizing}
L.~Van~der Maaten and G.~Hinton, ``Visualizing data using t-sne,'' \emph{Journal of Machine Learning Research}, vol.~9, no.~11, 2008.

\end{thebibliography}

\end{document}